\documentclass[]{style/style}
\usepackage{microtype}
\usepackage{graphicx}
\usepackage{subcaption}
\usepackage{csquotes}
\usepackage{afterpage}
\usepackage{xcolor}

\usepackage{amsmath}
\usepackage{amssymb}
\usepackage{mathtools}
\usepackage{amsthm}
\usepackage{xspace}
\usepackage{colortbl}
\usepackage{multirow}
\usepackage{multicol}
\usepackage{enumitem}
\usepackage{wrapfig}
\usepackage{nicefrac}
\usepackage[font=small,labelfont=bf,justification=justified,singlelinecheck=false]{caption}

\usepackage{booktabs}
\usepackage{multirow}
\usepackage{graphicx}
\usepackage{xcolor}
\usepackage{geometry}
\usepackage{array}
\usepackage{tabularx} % 用于自动计算列宽
\newcolumntype{L}[1]{>{\raggedright\arraybackslash}p{#1}}
\newcolumntype{Y}{>{\raggedright\arraybackslash}X}

\usepackage{amsmath}  
\newcommand{\model}{ReWorld\xspace}
\title{ReWorld: An Interactive World Model with Long-Horizon Memory}

\author[1,2,*]{Zhifei Chen}
\author[1,*]{Luozhou Wang}
\author[1,2]{Guibao Shen}
\author[1]{Dongyu Yan}
\author[1]{Shuai Yang}
\author[1]{Tianshuo Xu}
\author[1]{Yihua Du}
\author[2]{Wei Wang}
\author[2]{Tianyi Gui}
\author[2]{Lianghua Huang}
\author[1,\dagger]{Yingcong Chen}

\affiliation[1]{HKUST(GZ)}
\affiliation[2]{ATH, Alibaba}

\contribution[*]{Equal contribution}
\contribution[\dagger]{Corresponding author}

\renewcommand{\thefootnote}{\fnsymbol{footnote}}

\abstract{
An interactive world model must do three things at once: follow the user's actions, remember the places it has already shown, and keep streaming in real time. The tension is structural---control wants a short horizon, memory wants an unbounded one. ReWorld resolves it by separating the two during training and bounding them at inference. Mixed per-head attention windows confine most heads to the recent past while a small set of global heads attends over the entire history, and random head routing keeps either capability from binding to particular heads; random chunk dropping makes sparse histories in-distribution. At inference the whole past lives under a fixed budget: a bounded KV cache backed by a pose-indexed landmark bank, from which the model retrieves the landmarks nearest the current pose. A metric-scale-aligned data engine places eight sources---Unreal-rendered fly-throughs, game roaming, and real-world footage---on one physical action scale, so the same key press moves the camera the same distance in every source, and palindrome trajectories supply the revisit evidence that memory training needs. Distribution-matching distillation confined to a LoRA adapter then compresses sampling to four steps: one backbone serves both a high-fidelity multi-step mode and a real-time interactive one, streaming $704\times1280$ video across photorealistic, game-style, and stylized worlds. Under a three-axis protocol covering action following, long-horizon recall, and video quality, against six recent interactive world models it attains the best control fidelity ($11.95^\circ$ rotation error and the best camera-motion consistency) and the best generation quality; and on minute-long out-and-back rollouts ($64$\,s, $384$ latents), its fixed 12-chunk cache still regenerates the starting view---at rollout lengths where a sliding window has long evicted the evidence and full-KV attention runs out of memory.
}

\metadata[Website]{\url{https://zhifeichen097.github.io/ReWorld/}}
\teaserfig{%
  \begin{center}
  \includegraphics[width=\linewidth]{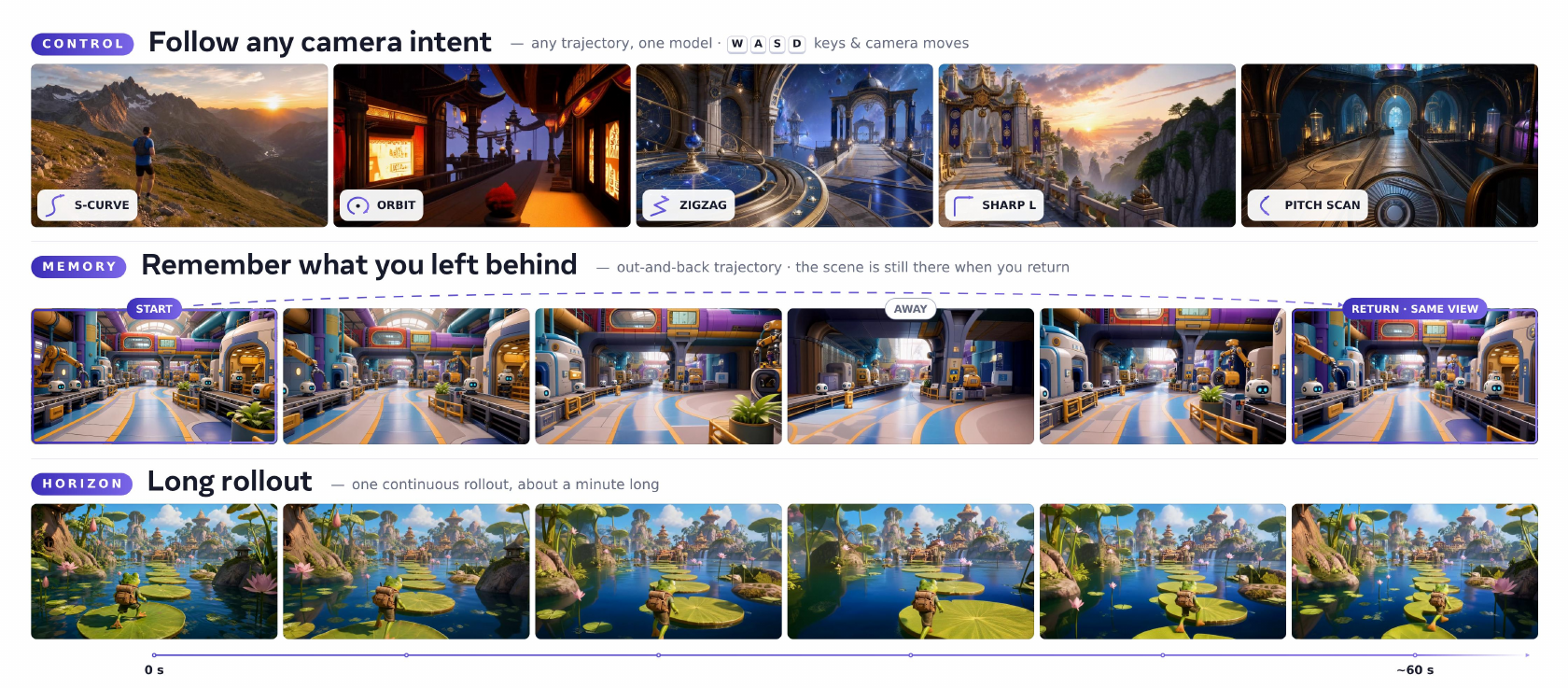}
  \captionof{figure}{\textbf{ReWorld} follows camera intents (top), regenerates a revisited view after an out-and-back excursion (middle), and streams long rollouts (bottom).}
  \label{fig:teaser}
  \end{center}%
}

\begin{document}

\maketitle
{\let\thefootnote\relax\footnotetext{The project was done during an internship at Alibaba.}}

\section{Introduction}
\label{section:intro}

A world model simulates an environment an agent can act in \citep{ha2018worldmodels}. Recent interactive systems turn a stream of user actions into video of a coherent, explorable environment \citep{bruce2024genie, genie3, worldplay2025, matrixgame2026, lingbotworld2026}. Expectations have converged: the model should react, reflecting a key press in the very next frames; remember, so that a revisited place looks the way it did before; and stream, generating at interactive rates over unbounded rollouts.

Current systems pursue these goals along two axes. For control, one route folds camera pose into attention, so attention logits depend on relative pose \citep{remind2026, dreamx2026}; the other directly injects an explicit action signal \citep{worldplay2025, matrixgame2026}. For memory, camera pose is the retrieval key of choice, in three forms: selecting which past frames re-enter the context \citep{yu2025context}, indexing an external memory bank \citep{worldmem, spatialmem2025}, or entering the attention index itself \citep{remind2026}.

Learning a world model comes down to learning two abilities: control, so that the next frames follow the current action, and memory, so that a revisited place looks the way it did before. When the two are trained together, control learns well but memory does not: adding direct action injection improves every control metric while revisit fidelity drops (Sec.~\ref{sec:exp-abl}, Table~\ref{tab:fusion}). The two abilities also ask for different attention windows. Control should not depend on the window size: at inference the model must respond to the current action correctly whatever window it is given, and what the response requires---the current scene and the current command---sits inside a short window. Memory is the opposite: retrieval can only be learned under a long window, because a model that cannot see the far past has nothing to retrieve from. The learning of the two abilities can therefore be split by window: control learned under short windows, memory under long ones.

This paper presents \model, an action-controllable streaming world model designed in two steps: split the training of control and memory by window, then consolidate memory under a fixed KV budget at inference. We keep both conditioning channels: pose-indexed attention (MRoPE) finds cached content by pose, and direct action injection states the command. What we separate is the window each attention head trains under. Mixed attention windows give the model both window lengths in every iteration: most heads are \emph{local} and attend only to a short recent window, while a small set of \emph{global} heads attends over the entire causal past. Random head routing then switches which heads are global at every step, cycling through a fixed pool of random partitions. The routing is what makes the split workable. The split itself cannot be kept at inference: the full history is gone, and all that remains is one bounded cache that every head reads, so there is no way to give some heads a long window and others a short one. A fixed partition would train heads to specialize in windows that deployment cannot provide. Routing removes this dependence: which heads are global changes at every step, every head trains under both windows, and neither ability binds to particular heads---at inference any head can read the shared bounded cache. The recipe adds no parameters, losses, or modules, and it makes a testable prediction: swapping cache-compression policies at inference should leave control error unchanged (Sec.~\ref{sec:exp-abl}).

With control insulated, the second step strengthens memory at inference: the rollout is unbounded but the cache is not. \model therefore \emph{consolidates}, keeping few chunks but keeping them intact: a chunk aging out of the recent window is stored in a bounded landmark bank only if the camera has travelled far enough since the last stored landmark---one full-resolution snapshot per stretch of camera travel---and once the bank is full, each admission evicts the member most spatially redundant with the rest. \emph{Retrieval} then fills the fixed cache with a sink chunk \citep{streamingllm}, a recent window, and the landmarks closest to the current camera pose. Inference thus reads a sparse, non-contiguous history, while standard training sees only a complete, contiguous prefix---a train--test mismatch \citep{huang2025selfforcing}. \emph{Chunk-drop training} closes the gap: a random subset of past chunks is masked at every step, teaching the model to reconstruct scene state from incomplete memory and making the spliced caches of deployment in-distribution (Secs.~\ref{sec:chunkdrop} and~\ref{sec:boundedkv}).

Two components complete the system. A metric-scale-aligned pipeline places synthetic, real, and game footage on a single physical action scale, so the same key press moves the camera the same physical distance in every source, with palindrome trajectories supplying the revisit evidence memory training needs (Sec.~\ref{sec:data}). Distribution-matching distillation with self-forcing rollouts \citep{yin2024onestepdiffusiondistributionmatching, yin2024improveddistributionmatchingdistillation, huang2025selfforcing} compresses sampling into a few denoising steps inside a LoRA adapter \citep{hu2022lora}, so one backbone serves both a high-fidelity multi-step operating point and a real-time interactive one (Sec.~\ref{sec:distill}).

In summary, our contributions are:
\begin{itemize}
    \item \textbf{Window-split training of control and memory.} Mixed per-head attention windows train control under short windows and memory under long ones, and random head routing keeps control in every head complete within the short window, so cache compression at deployment does not disturb action following (Secs.~\ref{sec:arch} and~\ref{sec:exp-abl}).
    \item \textbf{Chunk-drop training with consolidation-based inference.} A training augmentation that makes sparse, non-contiguous KV caches in-distribution, paired with an inference mechanism that consolidates aged chunks into a landmark bank kept bounded by redundancy-based eviction and retrieves them by pose proximity, so spatial memory persists over unbounded rollouts at a fixed KV budget (Secs.~\ref{sec:chunkdrop} and~\ref{sec:boundedkv}).
    \item \textbf{A metric-aligned multi-source data pipeline.} Synthetic, real, and game footage aligned to one physical action scale, with palindrome augmentation providing revisit supervision (Sec.~\ref{sec:data}).
    \item \textbf{Real-time deployment and evaluation.} LoRA-confined few-step distillation that gives a single backbone both a high-fidelity and a real-time operating point, together with an evaluation protocol covering action following, long-horizon recall, and general video quality (Secs.~\ref{sec:distill} and~\ref{sec:exp}).
\end{itemize}

\FloatBarrier

%%% Method (three parts merged into one section) %%%
\section{Method}
\label{sec:method}

\subsection{Overview}
\label{sec:overview}

ReWorld is an action-controllable streaming world model: given a text prompt, an optional reference image, and a stream of 6-DoF camera actions, a causal flow-matching diffusion transformer~\citep{lipman2023flow, esser2024scaling} generates video one latent chunk at a time, each chunk driven by a per-chunk camera action. The generated stream must both follow the commanded trajectory and remain spatially consistent with everything already generated---for instance, when the camera revisits a location seen long ago. The two requirements are learned under different conditions: action following needs only what a short window holds, while spatial memory can be learned only when distant content is visible in the attention window. Deployment adds a hard constraint: the KV cache holds a limited number of chunks, so the model cannot simply attend to its full history.

\begin{figure}[t]
\centering
% Source: figures/framework_fig.html --- edit it, then run: python figures/export_figs.py
\includegraphics[width=\linewidth]{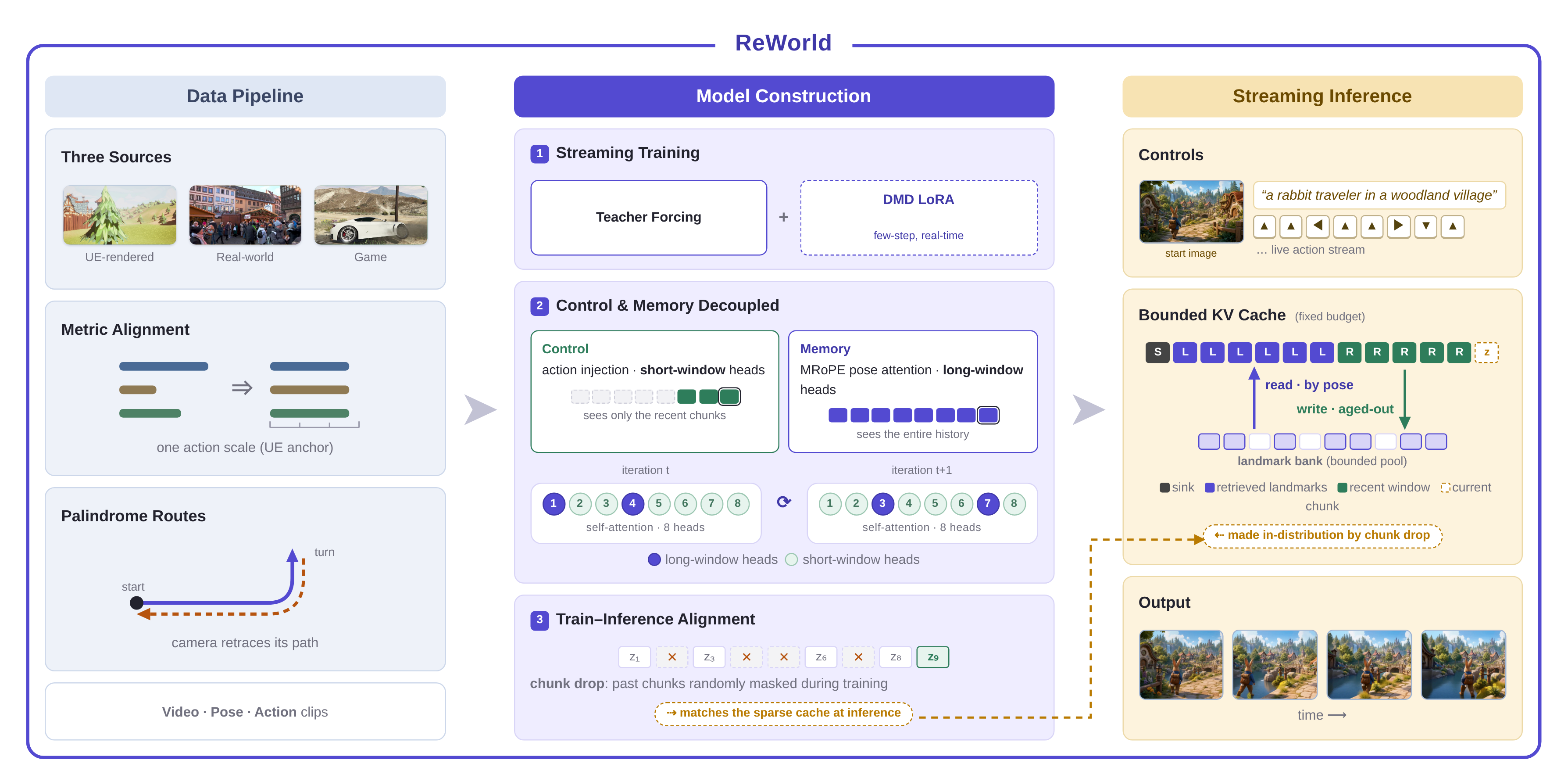}
\caption{\textbf{Overview of \model.} \emph{Left:} a metric-aligned data pipeline places UE-rendered, real-world, and game footage on a single physical action scale, and palindrome routes---the camera retracing its own path---supply revisit supervision (Sec.~\ref{sec:data}). \emph{Middle:} teacher-forcing training turns the bidirectional backbone into a streaming world model, with a DMD LoRA trained alongside for few-step real-time inference (Secs.~\ref{sec:arch} and~\ref{sec:distill}); control (action injection, local short-window heads) and memory (MRoPE pose-indexed attention, global long-window heads) are decoupled, and the partition into global and local heads is switched every step through a fixed pool of random partitions so that neither capability binds to particular heads (Sec.~\ref{sec:routing}); chunk-drop training masks random parts of the history, matching the sparse cache the model will read at inference (Sec.~\ref{sec:chunkdrop}). \emph{Right:} at inference, a fixed cache budget holds a sink chunk, pose-retrieved landmarks, and a recent window next to the chunk being generated; chunks aging out of the recent window are consolidated into a bounded landmark bank and retrieved by pose proximity upon revisits, so spatial memory persists over unbounded rollouts at constant cost (Sec.~\ref{sec:boundedkv}).}
\label{fig:framework}
\end{figure}

ReWorld meets both requirements with a small set of co-designed components. The camera-control section builds up from \textbf{pose-indexed attention (MRoPE)} to direct action injection, then trains control under short windows and memory under long ones with \textbf{mixed per-head attention windows} and \textbf{random head routing} (Sec.~\ref{sec:arch}); memory consolidation then strengthens what the decoupling protects---at inference a \textbf{pose-retrieved landmark cache} fills the fixed budget with the most relevant old chunks (Sec.~\ref{sec:boundedkv}), and chunk-drop training makes attention robust to the sparse caches this policy produces (Sec.~\ref{sec:chunkdrop}); \textbf{four-step LoRA distillation} makes the pipeline real time (Sec.~\ref{sec:distill})---the overall development route, AR-training the base model then plugging in a step-distilled LoRA, follows LongLive-2.0~\citep{longlive2026}; and a metric-aligned, eight-source data mixture lets a single action space transfer across synthetic and real footage (Sec.~\ref{sec:data}). Figure~\ref{fig:framework} summarizes the design.

\subsection{Camera-Controllable Streaming Generation}
\label{sec:arch}

\paragraph{Backbone and chunked causal generation.}
ReWorld builds on Wan2.2-TI2V-5B~\citep{wan2025wan}, a video diffusion transformer~\citep{peebles2023scalable} operating in the latent space of a causal VAE. We turn this bidirectional backbone into a streaming generator by imposing chunkwise causality: video is generated as a sequence of latent chunks ($L{=}12$ chunks of four latent frames per window), with full attention within a chunk and causal attention across chunks, so at inference the model emits one chunk per denoising pass and appends its keys and values to the cache $\mathcal{C}$.

\paragraph{Pose-indexed attention as implicit memory.}
Standard RoPE indexes attention by time and space, so a revisited location is, positionally, just a distant timestamp---retrieval must be inferred from content alone. Our first component therefore builds spatial memory into the attention index itself. We adopt Memory-RoPE (MRoPE), following the camera-phase RoPE design of PM-RoPE~\citep{remind2026}: each latent frame $f$ carries a relative camera-to-world pose $P_f \in \mathrm{SE}(3)$ (anchored to the first frame) with descriptor $c_f = [\mathrm{vec}(R_f);\, t_f] \in \mathbb{R}^{12}$, which a zero-initialized MLP maps to a phase offset $\delta_f$ applied on top of RoPE to queries and keys within the same attention pass,
\begin{equation}
\tilde{q} \;=\; \mathrm{RoPE}(q)\, e^{\,i\,\delta_{f(q)}}, \qquad
\tilde{k} \;=\; \mathrm{RoPE}(k)\, e^{\,i\,\delta_{f(k)}}, \qquad
\langle \tilde{q}, \tilde{k} \rangle \;\propto\; e^{\,i\,(\delta_{f(q)} - \delta_{f(k)})},
\label{eq:prope}
\end{equation}
so attention depends on pose difference rather than temporal distance: similar viewpoints are pulled together no matter how far apart in time. The cache thus acts as an implicit spatial memory---each chunk is stored with the pose it was seen from, and a revisit retrieves it by pose proximity. Zero-initialized $\mathrm{SE}(3)$ residuals on the values and outputs ($\tilde{v} = v + W_v(P_f^{-1} \circ v)$, $\mathrm{out} = W_o y + W_p(P_f \circ y)$) complete the conditioning. Unlike the two-pass designs in Table~\ref{tab:prope}, which add a separate camera-aware attention pass so that pose can \emph{steer} generation, MRoPE uses pose as a \emph{retrieval index} in the existing pass: the cost is one small MLP plus two linears, and the attention kernel and per-head windows stay untouched. \textbf{MRoPE alone gives strong revisit memory but imprecise control} (Sec.~\ref{sec:exp-abl}, Table~\ref{tab:fusion}).

\begin{table}[t]
\centering
\caption{Pose conditioning in attention: design comparison. Two-pass designs such as HY-World~1.5 add a camera-aware attention pass next to the temporal-RoPE pass and fuse the outputs; E-PRoPE trims the second pass to spatially reduced tokens; MRoPE folds pose into the same pass, leaving the attention kernel, mask layout, and per-head windows unchanged.}
\label{tab:prope}
\small
\resizebox{\linewidth}{!}{%
\begin{tabular}{@{}lllcl@{}}
\toprule
\rowcolor{stylegreen}
\textcolor{white}{\textbf{Design}} & \textcolor{white}{\textbf{Attention index}} & \textcolor{white}{\textbf{Where pose enters}} & \textcolor{white}{\textbf{Attn.\ passes}} & \textcolor{white}{\textbf{Extra cost}} \\
\midrule
Temporal RoPE & time \& space & --- & 1 & --- \\
HY-World 1.5 (two-pass)~\citep{worldplay2025} & time $+$ projective pose & separate camera-aware pass, fused & 2 & ${\approx}2\times$ attention FLOPs \\
E-PRoPE~\citep{dreamx2026} & time $+$ projective pose & separate reduced-token camera pass, added back & 2 & one reduced pass (${\sim}4.5\times$ fewer tokens) \\
MRoPE (ours) & time, space \& pose & Q/K phase $+$ zero-init V/O residuals & 1 & 1 small MLP $+$ 2 linears \\
\bottomrule
\end{tabular}%
}
\end{table}

\paragraph{Action injection.}
For control authority, we therefore inject the camera command directly. The commanded pose of each latent frame is expanded into a Pl\"ucker ray map: every spatial position receives the 6-D Pl\"ucker coordinates $[d,\; o \times d]$ of its viewing ray under that pose (direction $d$, camera center $o$, expressed in the same first-frame coordinate system as the MRoPE poses). An MLP projects the map to the model width, and the result is added token-wise to the patch embeddings at the transformer input (Figure~\ref{fig:framework}). The two channels now have clearly separated jobs: the ray map tells each token of the \emph{current} chunk where its camera should look, while MRoPE poses tell attention where every \emph{cached} token was seen. They still encode the same trajectory, however, and when every head receives both, the action signal crowds out pose-keyed retrieval: \textbf{control improves, but long-horizon memory degrades} (Sec.~\ref{sec:exp-abl}).

\begin{figure}[t]
\centering
% Source: figures/detail_figs.html --- edit it, then run: python figures/export_figs.py
\includegraphics[width=\linewidth]{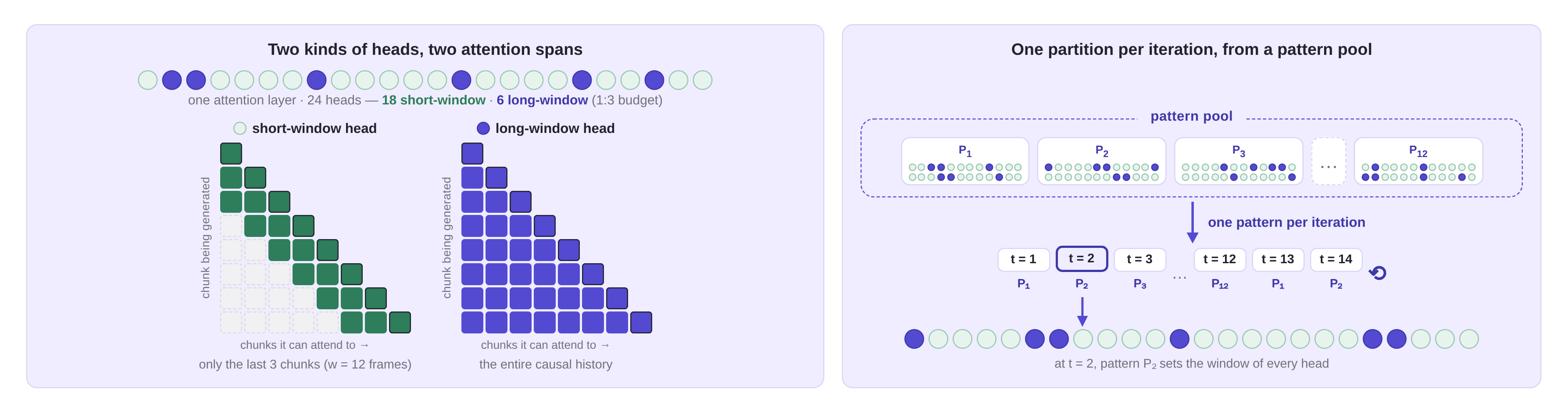}
\caption{\textbf{Mixed per-head attention windows, routed from a pattern pool.} \emph{Left:} local heads attend to the last $w{=}12$ frames (reactive control); global heads attend to the full causal history (long-range memory). \emph{Right:} each optimizer step switches to the next of $|\mathcal{P}|{=}12$ random head partitions, so neither ability binds to particular heads.}
\label{fig:routing}
\end{figure}

\paragraph{Mixed per-head attention windows.}
The two abilities need different windows to be learned. Action following does not depend on the window size: the next chunk must answer the current command however much history is visible, and the current scene plus the command fit in a short window. Retrieval, by contrast, can only be learned when distant content is visible. We therefore train under both windows at once (Fig.~\ref{fig:routing}, left): we split the $H{=}24$ attention heads of every block into 18 \emph{local} heads, which attend only to the last $w{=}12$ latent frames (three chunks), and a set $\mathcal{G}$ of $|\mathcal{G}|{=}6$ \emph{global} heads, which attend to the entire causal history:
\begin{equation}
\mathrm{context}(h) \;=\;
\begin{cases}
\text{full causal history,} & h \in \mathcal{G} \ \text{(6 global heads)},\\
\text{last } w{=}12 \text{ frames,} & \text{otherwise \ (18 local heads)}.
\end{cases}
\label{eq:perhead_mask}
\end{equation}
Every iteration thus trains both abilities at once: local heads learn control under the short window, and global heads are the only heads that can see distant content, so they are where retrieval is learned (a fixed $1{:}3$ global-to-local budget).

\paragraph{Random head routing.}\label{sec:routing}
A fixed partition would not survive deployment. At inference the per-head split is hard to realize: the full history is gone, and all heads read the same bounded cache (Sec.~\ref{sec:boundedkv}), so no head may be trained to depend on a particular window. We therefore draw a pool $\mathcal{P}$ of $|\mathcal{P}|{=}12$ random six-head global sets, and every optimizer step switches to the next set in the pool (Fig.~\ref{fig:routing}, right); the $1{:}3$ global-to-local ratio is unchanged. Every head thus spends steps in both roles: neither ability binds to particular heads, and since a head is local on most steps, the control it learns must be complete within the short window; retrieval keeps living in whichever heads currently see far. Changing the inference cache policy should then leave control error unchanged---a prediction Sec.~\ref{sec:exp-abl} tests.

This is the control-side design of \model: pose-indexed attention supplies the memory index, action injection supplies the command, and mixed-window training with random head routing lets each ability learn under the window it needs. Its memory-side counterpart, chunk-drop training and landmark retrieval, follows in the next sections.

% \paragraph{Training details.}
% ReWorld is trained in two stages. Stage one trains at 480p ($384\times 640$; $24\times 40$ latent grid) on the metric-aligned multi-source mixture (Sec.~\ref{sec:data}). Stage two warm-starts at 720p ($704\times 1280$; $44\times 80$ latents, 880 tokens per frame), where naively reusing 480p weights leaves cold rows in the spatial RoPE table and dilutes global-head softmax over the $3.67\times$ longer per-frame token count; we therefore rescale the spatial RoPE position axes by the ratio of the 480p to 720p post-patch grids (position interpolation on height and width only, leaving the temporal axis untouched), which preserves the spatial frequency range the model was trained on. Both stages use teacher-forcing diffusion training with timestep shift 5.0 over 1000 timesteps, classifier-free training guidance 3.0, AdamW with learning rate $10^{-5}$ and $(\beta_1, \beta_2) = (0.0, 0.999)$, EMA decay 0.99, and hybrid-sharded FSDP with sequence parallelism of degree 2. For real-time deployment, the trained model is further distilled into a 4-step generator with DMD-style self-forcing distillation~\citep{yin2024onestepdiffusiondistributionmatching, huang2025selfforcing} using LoRA adapters (rank 128); we defer these details to Sec.~\ref{sec:distill}.

\subsection{Memory Consolidation under a Bounded KV Budget}
\label{sec:memory}

ReWorld generates video autoregressively over latent chunks $z_k$, so its KV cache $\mathcal{C}$ grows linearly with the rollout horizon: at 720p, an unbounded cache exhausts device memory well before the horizons at which spatial memory is actually tested. A deployable interactive world model must therefore read from a cache of \emph{constant} size while still recalling scene content observed arbitrarily far in the past~\citep{yu2025context,worldplay2025,matrixgame2026,spatialmem2025}. ReWorld addresses this by separating the write side of memory from the read side. On the write side, \emph{consolidation} keeps few chunks and keeps them intact and diverse: aged chunks are admitted selectively into a bounded landmark bank at full resolution---one snapshot per stretch of camera travel---and the bank is held at its capacity by evicting the member most spatially redundant with the rest---rather than keeping the whole history at degraded fidelity. On the read side, \emph{retrieval} spends a fixed budget of $B{=}12$ chunks on a pose-indexed working set. A training-time augmentation, chunk drop, makes the backbone robust to the sparse caches this policy produces. The pose-indexed positional encoding of Sec.~\ref{sec:arch} is what makes retrieval effective: because attention keys carry camera-pose structure, a retrieved chunk remains addressable by pose regardless of how distant it is in time.

\begin{figure}[t]
\centering
% Source: figures/detail_figs.html --- edit it, then run: python figures/export_figs.py
\includegraphics[width=\linewidth]{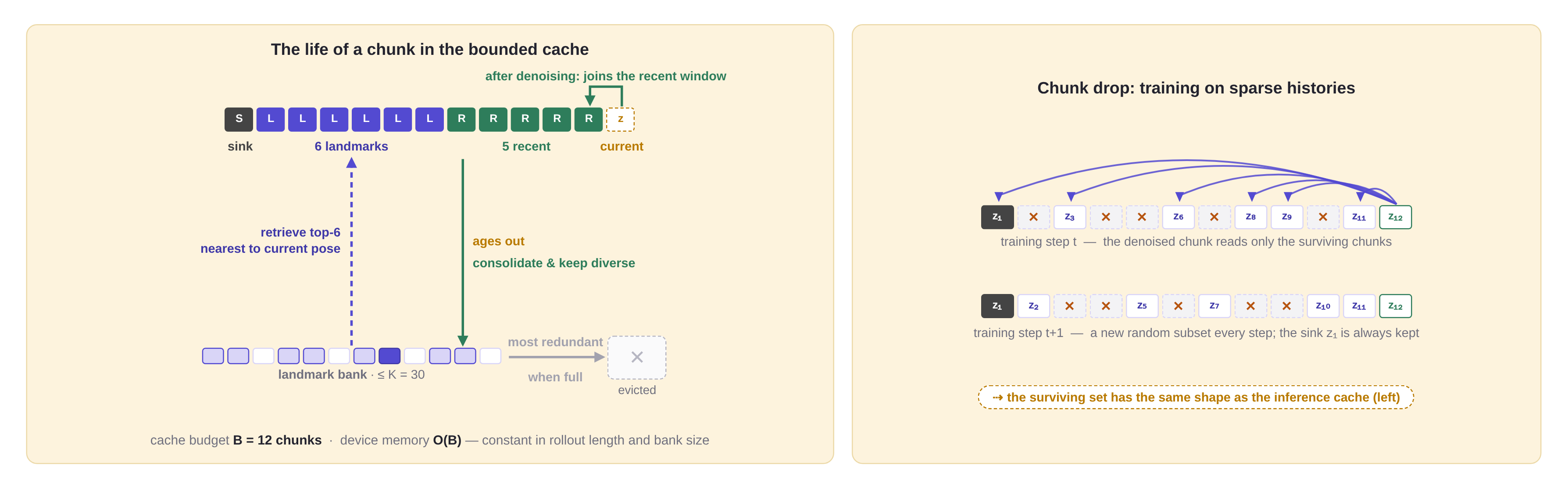}
\caption{\textbf{Memory consolidation under a fixed KV budget.} \emph{Left:} at inference the cache holds one sink, six pose-retrieved landmarks, and five recent chunks ($B{=}12$); a chunk aging out of the recent window is consolidated into the bounded landmark bank---kept sparse and diverse---and re-enters the cache when the camera returns nearby (a full bank evicts its most redundant member). \emph{Right:} chunk-drop training randomly masks past chunks at every step, so such sparse caches are in-distribution.}
\label{fig:consolidation}
\end{figure}

\paragraph{Landmark consolidation and retrieval.}
\label{sec:boundedkv}
At deployment (Fig.~\ref{fig:consolidation}, left), ReWorld maintains a bounded cache whose active set at chunk $k$ is
\begin{equation}
\label{eq:cache}
\mathcal{C}_k \;=\; \underbrace{\{\,1\,\}}_{\text{sink}} \;\cup\; \underbrace{\{\,k{-}5,\dots,k{-}1\,\}}_{\text{recent window}} \;\cup\; \underbrace{\mathcal{T}_k}_{\text{retrieved landmarks}},
\qquad |\mathcal{T}_k| = 6,\quad |\mathcal{C}_k| = B = 12 ,
\end{equation}
where the sink chunk anchors the global scene layout in the spirit of attention sinks~\citep{streamingllm}, and every retained chunk is stored at full resolution---no pooling, merging, or token-level pruning is applied. The retrieved landmarks are drawn from a bounded bank $\mathcal{M}$ ($|\mathcal{M}| \le K$, $K{=}30$) held in a tiered store: a pinned host-memory master copy of all $K$ members, with only the retrieved working set staged on device. Each chunk carries a single camera pose $P_k \in \mathrm{SE}(3)$ (first-frame rotation with block-mean translation), and the bank is managed by pose alone. Admission keeps the bank diverse with an \emph{odometer} rule: a chunk aging out of the recent window is stored only if the camera has travelled at least a stride $\delta$ (measured in median step lengths) since the \emph{last stored} landmark; the chunks in between are the densest, most redundant samples along a path, and are dropped. Crucially, the rule measures distance travelled---never whether a place is already banked---so a revisit pass is stored just like the first visit, whereas gating against the whole bank would discard exactly the chunk a later revisit needs (only ${\sim}64\%$ far-revisit coverage in simulation). Once the bank is full, each admission evicts the member most redundant in pose space (the one whose nearest neighbour is closest), and the first two members are never evicted, since the earliest region is the one revisited across the longest gaps. After every generated chunk, the six landmarks nearest to the current pose re-enter the cache as $\mathcal{T}_k$, where they are read through pose-indexed attention. Device memory thus stays at $O(B)$ chunks regardless of rollout length, with the full bank in host memory; landmark transfers are prefetched on a separate stream and overlap with denoising.

% Crucially, the budget is counted in \emph{chunks} (equivalently frames), not tokens: the per-frame token count scales with spatial resolution ($3.67\times$ from 480p to 720p under our latent grids), so a token-denominated budget would silently retain $3.67\times$ fewer frames at 720p and evict distant observations earlier at higher resolution. Chunk-denominated budgets keep the retained temporal horizon resolution-invariant, which we found essential for fair cross-resolution comparison.

\paragraph{Chunk-drop training.}
\label{sec:chunkdrop}
The cache above hands the model a sparse, non-contiguous subset of its history at inference, whereas standard teacher forcing conditions every denoising query on a \emph{contiguous, complete} causal prefix---a train--test gap that manifests as blur and drift as soon as chunks are evicted. We close it by randomly dropping cached KV chunks during training (Fig.~\ref{fig:consolidation}, right): for a training clip of $L{=}12$ chunks (Sec.~\ref{sec:overview}), the sink chunk is always kept and five further survivors are drawn at random at every optimization step, so each step exposes the model to a random half of its history ($6$ of $12$ chunks). The attention mask is restricted accordingly: the query for chunk $z_k$ attends only to the clean keys of surviving earlier chunks and to its own chunk, and queries are never dropped, meaning that every chunk still receives its denoising loss. The keep-set is resampled independently at every step and broadcast so that all data- and sequence-parallel ranks apply an identical mask.

We compare this deployment policy against four alternatives under the identical budget $B{=}12$: an unbounded full-KV cache (undeployable upper bound), a sliding-window cache with sink (forgetting baseline), mean-pooled compression of aged chunks, and a retrieval-free variant that replaces pose retrieval with a static bank of six pose-deduplicated landmarks attended in full; results are reported in Sec.~\ref{sec:exp-abl} (Tab.~\ref{tab:ablation}).

\subsection{Real-Time Distillation with a Lightweight Adapter}
\label{sec:distill}

\begin{wrapfigure}{r}{0.42\textwidth}
\centering
\includegraphics[width=\linewidth]{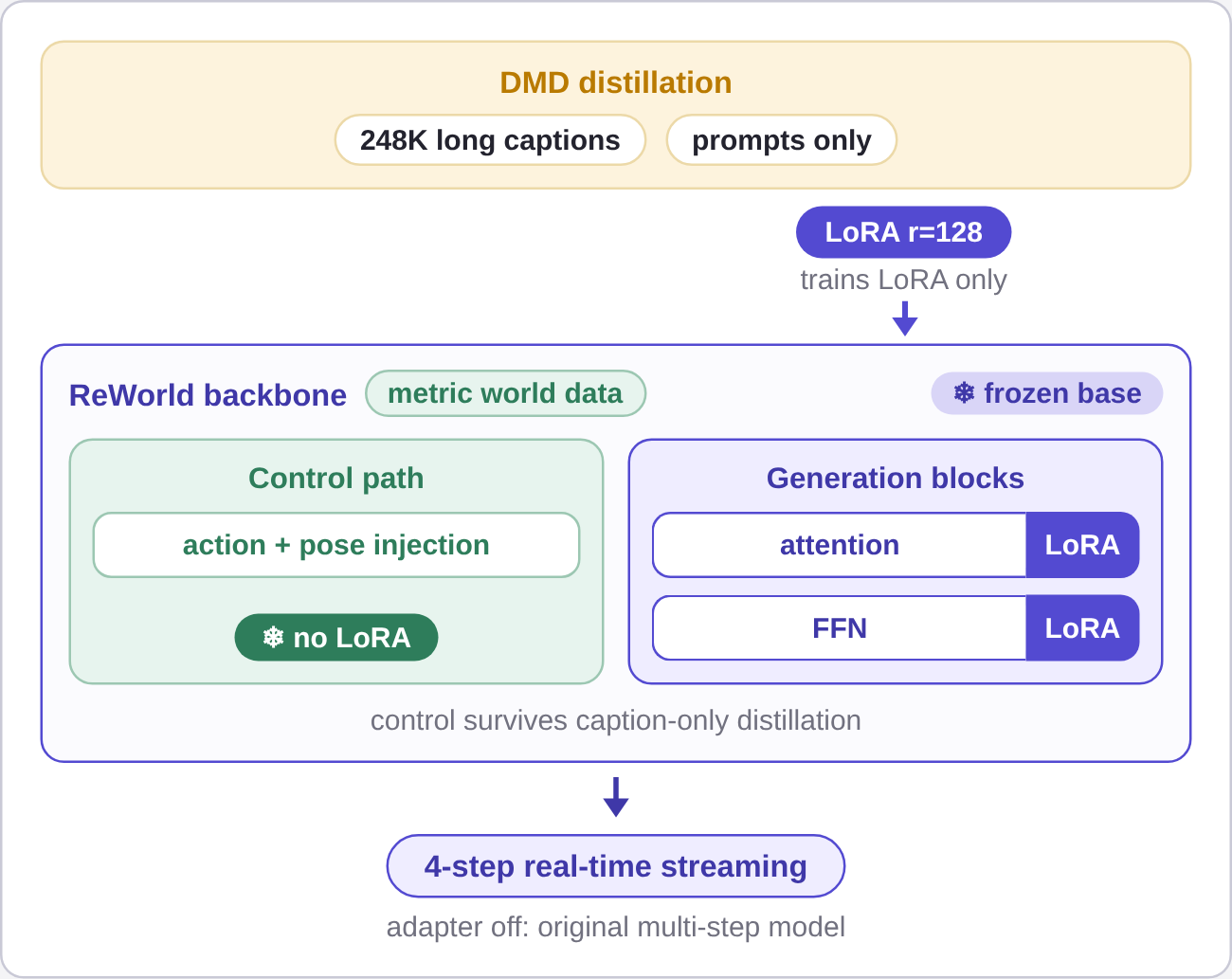}
\caption{\textbf{Real-time distillation.} A caption-domain DMD LoRA plugs into the frozen backbone; the control path carries no LoRA.}
\label{fig:dmd}
\end{wrapfigure}
To run in real time, ReWorld generates with only four denoising steps, obtained by step distillation confined to a plug-in LoRA in the AR-train-then-distill route of the LongLive series~\citep{longlive2025,longlive2026} (Fig.~\ref{fig:dmd}): the multi-step model is the teacher, and the student is the \emph{same backbone, frozen}, plus a rank-$128$ LoRA on the attention and feed-forward layers---the LoRA is the only thing trained (the DMD critic is also a LoRA over the frozen weights). Training follows distribution-matching distillation with self-forcing rollouts~\citep{huang2025selfforcing,yin2024onestepdiffusiondistributionmatching,yin2024improveddistributionmatchingdistillation}: the student streams chunks from its own KV cache, just as at deployment, so distillation needs only text prompts---no video. The prompts are generic long captions from another domain (248K, extended from VidProM), and the action and pose injection path carries no LoRA, so the model's control is untouched by distillation.

One backbone thus has two modes: adapter off is the original multi-step model; adapter on streams in real time at four steps. The 2.6\,GB adapter also transfers across base checkpoints, re-distilling in 2k--4k steps when needed. Two practical rules: distill at the deployment resolution, with prompts that resemble deployment prompts; the production student uses block size 16 with a single rollout and folds classifier-free guidance into the adapter.

\FloatBarrier

%%% Data pipeline %%%
\section{Data Pipeline}
\label{sec:data}

\begin{figure}[t]
\centering
% Source: figures/datapipeline_fig.html --- edit it, then run: python figures/export_figs.py
\includegraphics[width=\linewidth]{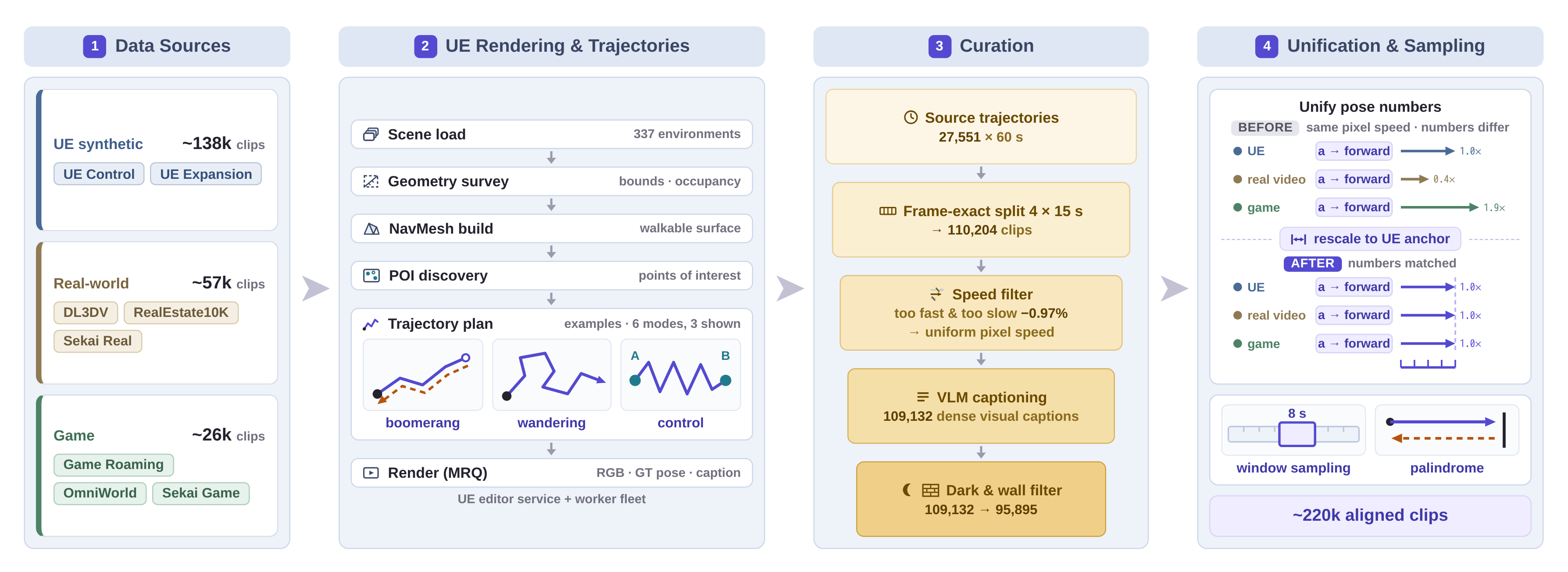}
\caption{\textbf{The four-stage data pipeline.} Eight UE, real-world, and game sources; an automated UE service renders $337$ environments with NavMesh-based trajectory planning (the control pattern anchors each path at two points of interest and fills the segment between them with direction-balanced random motion); a curation funnel filters each pool and evens out on-screen speed by dropping too-fast and too-slow clips; unification then rescales the pose numbers to one shared metric scale --- the same action moves the camera the same distance --- before clip-level sampling.}
\label{fig:datapipeline}
\end{figure}

ReWorld is trained on an eight-source joint corpus of $220{,}724$ pose-annotated clips (Fig.~\ref{fig:datapipeline}). Two of the sources are rendered by our own Unreal Engine (UE) pipeline and play a double role: they anchor the metric scale to which all other sources are aligned, and their trajectories are generated specifically for camera control.

\subsection{Sources}
\label{sec:data-sources}

Table~\ref{tab:data} summarizes the eight sources; each clip is stored as an RGB video, a per-frame camera trajectory, and a text caption. The sources fall into three groups --- UE-rendered (a control set and an expansion set, produced by the pipeline of Sec.~\ref{sec:data-ue}), real-world footage (DL3DV \citep{dl3dv}, RealEstate10K \citep{realestate10k}, and Sekai real-walking-hq \citep{sekai}), and game footage (game roaming, OmniWorld-Game \citep{omniworld}, and Sekai game-walking) --- with per-source construction details in Appendix~\ref{app:data}.

\begin{table}[t]
\centering
\caption{The eight-source joint training corpus. \#Clips are counts after per-source filtering (Sec.~\ref{sec:data-sampling}); Duration is the per-clip length; Pose gives the camera-trajectory provenance; Caption gives the text-annotation provenance.}
\label{tab:data}
\begin{tabular}{lrlll}
\toprule
\rowcolor{stylegreen}
\textcolor{white}{\textbf{Source}} & \textcolor{white}{\textbf{\#Clips}} & \textcolor{white}{\textbf{Duration}} & \textcolor{white}{\textbf{Pose}} & \textcolor{white}{\textbf{Caption}} \\
\midrule
UE control (ours) & $95{,}895$ & $15$\,s & engine GT (c2w, cm) & pipeline-generated \\
UE expansion (ours) & $42{,}536$ & $30$--$60$\,s & engine GT (c2w, cm) & scene summary + camera \\
DL3DV & $29{,}864$ & $15$\,s & VIPE & Qwen-VL \\
RealEstate10K & $12{,}065$ & $\geq 8$\,s & VIPE (re-estimated) & pre-computed \\
Sekai real-walking-hq & $14{,}730$ & $60$\,s & MegaSaM (c2w) & official CSV \\
Game roaming ($79$ games) & $18{,}387$ & $15$\,s & VIPE (metric) & Qwen-VL \\
OmniWorld-Game & $5{,}629$ & $\sim$$12$\,s & engine (w2c$\to$c2w) & per-clip JSON \\
Sekai game-walking & $1{,}618$ & $60$\,s & engine GT (UE5, c2w) & official CSV \\
\midrule
Total & $220{,}724$ & & & \\
\bottomrule
\end{tabular}
\end{table}

\subsection{UE Rendering Pipeline}
\label{sec:data-ue}

An automated service inside the UE editor drives scene loading, geometry survey, navigation-mesh construction, point-of-interest discovery, trajectory planning, and Movie Render Queue rendering, and a batch orchestrator sweeps this chain over a library of $337$ environments (Fig.~\ref{fig:datapipeline}). Each clip is exported with RGB frames, exact engine ground-truth camera poses (pinhole intrinsics and per-frame camera-to-world matrices), and scene and trajectory metadata; control clips additionally carry per-frame action labels.

Trajectories are planned for control, not only for scene coverage: they are drawn from six planned modes, each walking at a constant, mode-specific speed. The control mode anchors each trajectory at two points of interest and fills the path between them with randomized motion segments, so that backward, strafing, and diagonal motion are sampled as evenly as forward motion (Fig.~\ref{fig:datapipeline}). \textbf{The action-following accuracy reported in Sec.~\ref{sec:exp} owes as much to these direction-balanced trajectories as to the model design}, since forward-biased footage gives little supervision for the rarer directions.

\subsection{Metric-Scale Alignment}
\label{sec:data-metric}

The action $a_k \in \mathbb{R}^6$ that conditions each latent chunk is a frame-to-frame 6-DoF camera increment, so the same commanded action must correspond to the same motion in every source --- both on screen and in the pose numbers (Fig.~\ref{fig:datapipeline}). The curation stage already settles the on-screen half: its speed filter drops clips that move too fast or too slow, leaving the survivors at a similar pixel speed (Sec.~\ref{sec:data-sampling}). What still disagrees is the numbers attached to that motion --- estimated poses (VIPE, MegaSaM) are defined only up to scale, and engine poses differ in units. We therefore rescale the translations of each source $s$ by a single divisor --- the ratio of pooled per-latent-step translation-increment medians against the UE anchor, measured by replaying the exact training-time windowing with augmentation disabled:
\begin{equation}
\label{eq:transscale}
\sigma_s \;=\; \frac{\operatorname{med}_{s}\!\left[\lVert \Delta t \rVert\right]}{\operatorname{med}_{\mathrm{UE}}\!\left[\lVert \Delta t \rVert\right]},
\qquad
\operatorname{med}_{\mathrm{UE}}\!\left[\lVert \Delta t \rVert\right] = 0.3667,
\qquad
t \;\leftarrow\; t / \sigma_s .
\end{equation}
A single global scalar per source suffices because the per-clip scale spread within each source is bounded; all sources are further unified to camera-to-world matrices in a common camera basis, and per-source divisors, spread statistics, alignment audits, and convention verification are given in Appendix~\ref{app:data}.

\subsection{Filtering and Sampling}
\label{sec:data-sampling}

\paragraph{Filtering.}
The UE control renders pass the funnel of Fig.~\ref{fig:datapipeline} --- splitting, a speed filter that drops clips moving too fast or too slow, captioning, and a darkness/wall filter --- reducing $110{,}204$ clips to $95{,}895$, while the expansion set is drawn from the raw render pool by scene round-robin and darkness-filtered from $47{,}726$ to $42{,}536$ clips. The game-roaming source is curated from $494.7$ hours of gameplay across $168$ games down to $18{,}387$ free-roaming clips from $79$ games. Filters for the remaining sources are simple duration and completeness checks; all thresholds and per-source details are given in Appendix~\ref{app:data}.

\paragraph{Sampling.}
The eight sources are concatenated and sampled uniformly at the clip level, so the training mixture equals the clip-count proportions of Table~\ref{tab:data} --- $63\%$ UE, $26\%$ real, and $11\%$ game --- independent of clip duration. Each dataset access draws a random $189$-frame window resampled to $24$\,fps ($\approx 8$\,s, one training window; Sec.~\ref{sec:overview}); with probability $0.2$ on the two UE sources the window is replaced by a palindrome --- a random half of the window concatenated with its temporal reverse --- injecting the explicit revisit evidence that long-horizon spatial memory requires. From each window the loader emits the RGB frames, the per-chunk 6-DoF actions $a_k$, and the relative SE(3) trajectory $P_k$ consumed by MRoPE (Appendix~\ref{app:data}).

\FloatBarrier

%%% Experiments %%%
\section{Experiments}
\label{sec:exp}

\subsection{Setup}
\label{sec:exp-setup}

\paragraph{Implementation.}
ReWorld is trained in two resolution stages on top of the Wan2.2-TI2V-5B backbone~\citep{wan2025wan}: a 480p ($384\times640$) pre-training stage, followed by a 720p ($704\times1280$) warm-start with interpolated spatial RoPE positions; optimizer, guidance, and parallelism settings follow the training details of Sec.~\ref{sec:arch}. Each training window covers $L{=}12$ latent chunks $z_k$, conditioned on per-chunk actions $a_k$ and poses $P_k$ as defined in Sec.~\ref{sec:overview}. Of the $H{=}24$ attention heads, the global set $\mathcal{G}$ ($|\mathcal{G}|{=}6$) attends over the full causal history while the remaining heads use a local window of $w{=}12$ frames; during training, $\mathcal{G}$ cycles through a fixed pool of 12 random six-head partitions, switching every optimizer step (random head routing, Sec.~\ref{sec:routing}), and KV chunks are randomly dropped down to 6 kept chunks plus one sink chunk (chunk drop, Sec.~\ref{sec:chunkdrop}).

\paragraph{Real-time inference.}
Unless otherwise stated, all deployed and timed results are produced with a 4-step DMD-distilled~\citep{yin2024onestepdiffusiondistributionmatching, yin2024improveddistributionmatchingdistillation} LoRA (rank 128, block size 16, single-rollout student) applied to the EMA weights of the multi-step model, following the self-forcing style distillation recipe~\citep{huang2025selfforcing}. Inference runs at 720p with 4 denoising steps, CFG scale 1, and 16 latent frames per block; per-head attention windows are a training-time construct and are not applied at evaluation---all heads attend over the bounded cache. Bounded-memory arms operate under a KV-cache chunk budget of $B{=}12$ on the cache $\mathcal{C}$; the landmark bank $\mathcal{M}$ is capped at $|\mathcal{M}| \le K$ with $K{=}30$ (Sec.~\ref{sec:boundedkv}).

\paragraph{Evaluation suite.}
We evaluate four axes: (i) \emph{camera controllability} against six interactive world-model and camera-controlled video generation baselines on a shared trajectory benchmark (Sec.~\ref{sec:exp-camctrl}); (ii) \emph{long-horizon memory} with a needle-in-a-haystack (NIAH) protocol built from palindromic revisit trajectories of up to 384 latents (Sec.~\ref{sec:exp-niah}); (iii) \emph{video quality} on seven video-intrinsic VBench~\citep{vbench} dimensions (Sec.~\ref{sec:exp-vbench}); and (iv) \emph{ablations} isolating the training recipe (chunk drop, random head routing), the inference-time cache policy, and the action/pose fusion design (Sec.~\ref{sec:exp-abl}).

\subsection{Camera Controllability}
\label{sec:exp-camctrl}

\paragraph{Protocol.}
We construct a controllability benchmark of 40 start images (drawn from the six baselines' official repositories at $704\times1280$, so that no single method is favored) $\times$ 6 canonical trajectories, i.e., 240 clips per method. The trajectories are \emph{dolly} (sustained forward), \emph{strafe} (lateral left--right), \emph{arc\_yaw} (forward with yaw), \emph{arc\_pitch} (forward with pitch), \emph{s\_curve} (forward with alternating yaw), and \emph{palindrome} (forward then exact return); every trajectory contains forward translation, since pure-rotation ground truth has zero displacement and degenerates the similarity alignment. Each method receives the same camera intent translated into its native control interface (action keys, pose sequences, or text, as appropriate). Generated clips are re-tracked with ViPE~\citep{vipe}, and the estimated trajectory is registered to the intended trajectory by a $\mathrm{Sim}(3)$ position alignment together with a global orientation alignment (restricted to $\det = +1$), which absorbs per-method conventions in step magnitude, handedness, and coordinate frame so that the comparison measures whether the trajectory \emph{shape} follows the intent. We report per-trajectory rotation error RotErr (geodesic, degrees) and aggregate translation error TransErr and camera motion consistency CamMC. We compare against SANA-WM~\citep{sanawm2026}, DreamX~\citep{dreamx2026}, HY-WorldPlay~\citep{worldplay2025} (its lightweight Wan-based variant on this benchmark), Matrix-Game 3.0~\citep{matrixgame2026}, LingBot-World~\citep{lingbotworld2026}, and Yume-1.5~\citep{yume15}.

\begin{table}[t]
\centering
\caption{Camera controllability on the 40-image $\times$ 6-trajectory benchmark (240 clips per method). Overall RotErr, TransErr, and CamMC ($\downarrow$) averaged over all trajectories, followed by per-trajectory RotErr ($^\circ$, $\downarrow$) after $\mathrm{Sim}(3)$ and global orientation alignment. Best per column in bold.}
\label{tab:camctrl}
\resizebox{\linewidth}{!}{%
\begin{tabular}{l ccc cccccc}
\toprule
& \multicolumn{3}{c}{Overall} & \multicolumn{6}{c}{RotErr$^\circ$ $\downarrow$ per trajectory} \\
\cmidrule(lr){2-4} \cmidrule(lr){5-10}
\rowcolor{stylegreen}
\textcolor{white}{\textbf{Method}} & \textcolor{white}{\textbf{RotErr$^\circ$ $\downarrow$}} & \textcolor{white}{\textbf{TransErr $\downarrow$}} & \textcolor{white}{\textbf{CamMC $\downarrow$}} & \textcolor{white}{\textbf{dolly}} & \textcolor{white}{\textbf{strafe}} & \textcolor{white}{\textbf{arc\_yaw}} & \textcolor{white}{\textbf{arc\_pitch}} & \textcolor{white}{\textbf{s\_curve}} & \textcolor{white}{\textbf{palindrome}} \\
\midrule
SANA-WM~\citep{sanawm2026} & 13.02 & 0.123 & 0.388 & 4.95 & 1.51 & 32.01 & 21.33 & 17.54 & 0.80 \\
DreamX~\citep{dreamx2026} & 13.10 & 0.114 & 0.381 & 2.50 & 1.21 & 33.00 & 20.71 & 18.06 & 3.13 \\
HY-WorldPlay~\citep{worldplay2025} & 14.66 & 0.114 & 0.427 & \textbf{0.26} & \textbf{0.24} & 43.53 & 22.00 & 21.73 & \textbf{0.20} \\
Matrix-Game 3.0~\citep{matrixgame2026} & 15.45 & \textbf{0.075} & 0.394 & 8.19 & 1.00 & 39.93 & \textbf{20.59} & 17.75 & 5.24 \\
LingBot-World~\citep{lingbotworld2026} & 12.59 & 0.107 & 0.354 & 3.10 & 6.01 & \textbf{26.21} & 26.83 & \textbf{12.23} & 1.16 \\
Yume-1.5~\citep{yume15} & 14.24 & 0.131 & 0.428 & 0.84 & 2.01 & 29.05 & 24.57 & 25.60 & 3.38 \\
\midrule
\textbf{ReWorld (ours)} & \textbf{11.95} & 0.102 & \textbf{0.332} & 2.69 & 1.07 & 27.63 & 23.61 & 16.06 & 0.64 \\
\bottomrule
\end{tabular}}
\end{table}

Table~\ref{tab:camctrl} reports per-trajectory rotation error together with aggregate metrics for all seven methods. Because all methods are aligned with the same $\mathrm{Sim}(3)$ and orientation registration against a method-agnostic ground-truth intent, the comparison isolates trajectory-following fidelity from per-method magnitude conventions. ReWorld attains the best overall RotErr and CamMC, sits in the first tier on the translation-dominant trajectories (dolly, strafe, palindrome), and remains competitive on the rotation-heavy arcs, which are the hardest regime for every method.
% Caveats for the camera-ready pass: (1) RotErr is an absolute-angle metric, so it partly rewards methods that execute the commanded rotation magnitude; under a magnitude-normalized RotErr (align estimated rotation amount to GT, compare direction/shape only), SANA-WM/DreamX/LingBot-World come out slightly ahead -- i.e., we win by "turning enough", pure steering-shape precision still has headroom. (2) HY-WorldPlay here is its lightweight Wan-based variant (1280x704, 29 frames).

\subsection{Long-Horizon Memory}
\label{sec:exp-niah}

\begin{figure}[t]
\centering
\includegraphics[width=\linewidth]{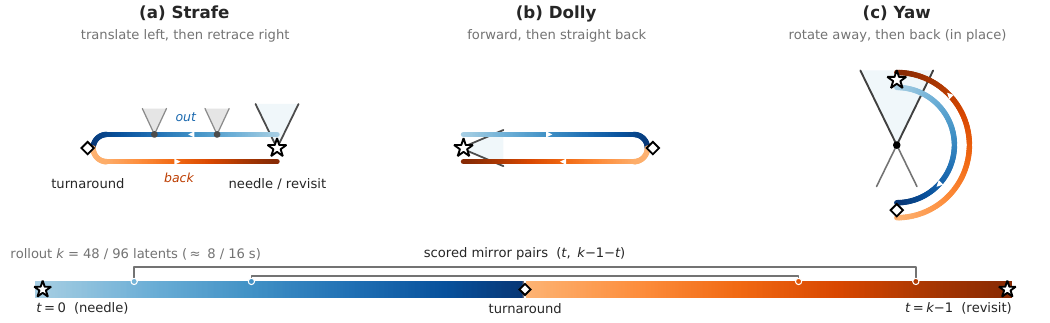}
\caption{\textbf{Constructed palindrome trajectories.} The camera moves out (blue) and retraces its path (orange), so the needle view at $t{=}0$ must be regenerated at $t{=}k{-}1$; scored mirror pairs $(t,\,k{-}1{-}t)$ span at least half the rollout.}
\label{fig:niah_traj}
\end{figure}

\paragraph{Protocol.}
We probe spatial memory with a needle-in-a-haystack protocol built on \emph{palindrome} trajectories: the camera moves out and then retraces its path, so views generated early in the clip must be reproduced after a long temporal gap---these early views are the needles. The baseline benchmark pairs 12 start images with 3 canonical palindromes at two rollout lengths $k \in \{48, 96\}$ latents ($\approx$8\,s and $\approx$16\,s), giving 36 clips per method and length. The palindromes are \emph{strafe}, \emph{dolly}, and \emph{yaw} (Fig.~\ref{fig:niah_traj}). Every method receives the same camera intent through its native control interface, with per-image matched prompts and a fixed seed. Scored revisit pairs are the mirror pairs of the palindrome, anchored in the earliest fifth of the clip and kept only when their gap spans at least half the rollout, up to five pairs per clip. For our own arms we additionally run a longer version of the memory test: grouped-action explorations completed into palindromes, at nested lengths $k \in \{96, 192, 288, 384\}$ latents; each length is a prefix of the same trajectory, so scores are comparable across $k$. This longer test feeds the ablations of Sec.~\ref{sec:exp-abl}.

\paragraph{Memory metrics.}
For each revisit pair we compare the predicted frame at the needle with the predicted frame generated when the camera returns to that pose, and score their similarity four ways. \emph{SSIM} measures pixel-level structure. \emph{LPIPS} is a perceptual distance between deep features (lower is better). \emph{DINO} is the cosine similarity of DINO ViT features and asks whether the two frames show the same scene layout. \emph{ORB} is the fraction of matched ORB keypoints and asks whether specific landmarks reappear in place. Each score is averaged over revisit pairs, then over clips, and reported as an absolute value. A rollout that barely moves scores high on any similarity metric, so the table also reports each method's executed \emph{path length}: the accumulated median optical flow of the outbound half, i.e., how much visual ground the rollout actually covers before turning back.

\begin{figure}[t]
\centering
\includegraphics[width=\linewidth]{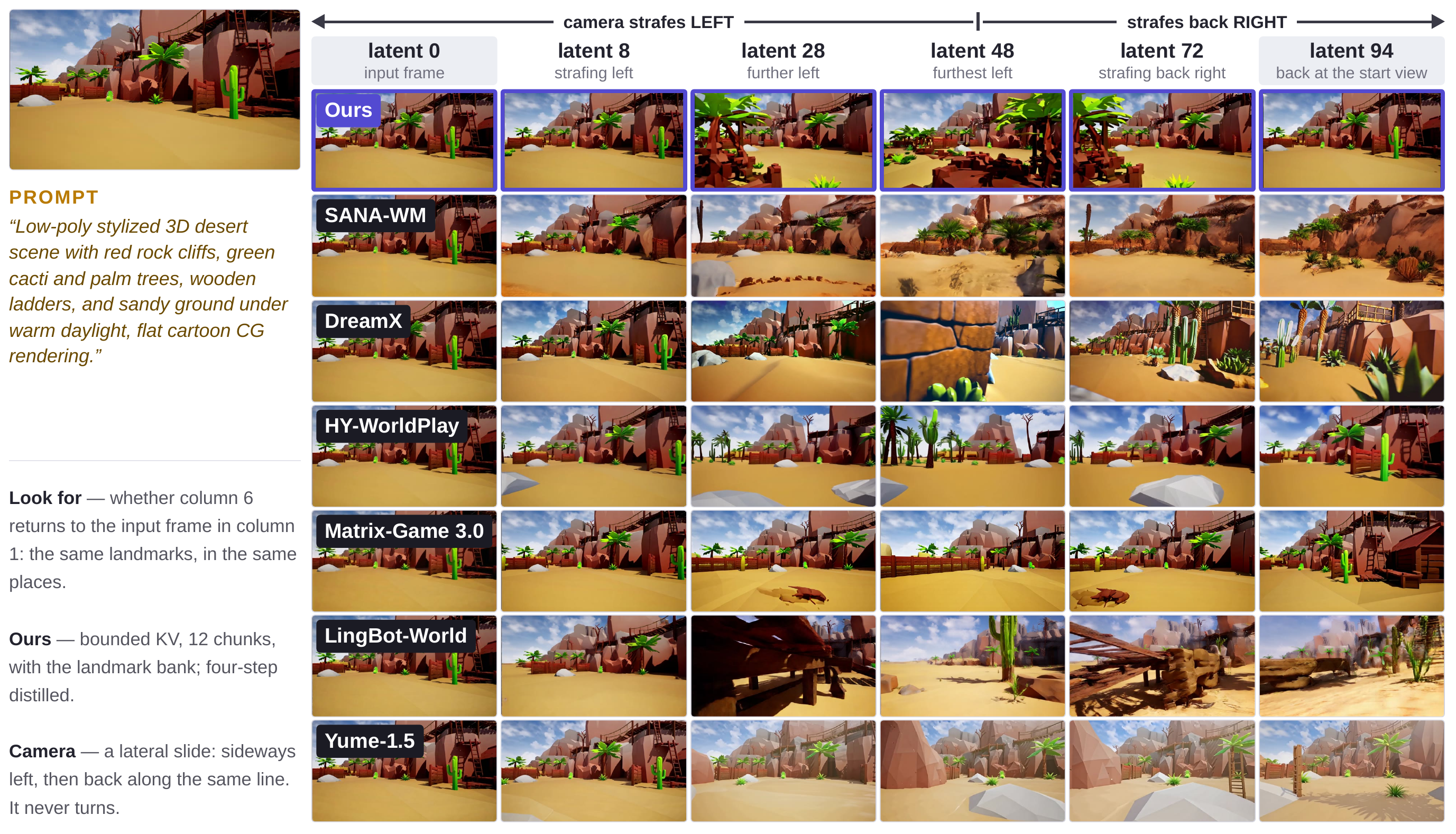}
\caption{\textbf{Qualitative memory comparison on a strafe-and-return rollout.} All methods start from the same input frame and execute the same lateral palindrome---the camera strafes left, then retraces the same line back to the start, without ever turning---so the last column should reproduce the first: the same landmarks, in the same places. Columns are latents $0/8/28/48/72/94$ of a $k{=}96$ rollout; each baseline is driven through its native control interface. ReWorld (top row; bounded cache $B{=}12$ with the landmark bank, four-step distilled) returns to the input view with the layout intact, whereas the baselines either drift off the commanded path or regenerate the revisited region with altered geometry.}
\label{fig:baseline_qual}
\end{figure}

\paragraph{Comparison to baselines.}
Table~\ref{tab:memory_base} reports all four scores at both lengths for the same six methods as Sec.~\ref{sec:exp-camctrl}. For HY-WorldPlay we evaluate its flagship autoregressive model, as in Sec.~\ref{sec:exp-vbench}. ReWorld runs its default deployment configuration (consolidation with the landmark bank, $B{=}12$). The two lengths play different roles. At $k{=}48$ the revisit gap reaches 47 latents, and several baselines still hold the start view in context---LingBot-World keeps a 52-latent window, while SANA-WM (linear attention) and Yume-1.5 are unbounded---so $k{=}48$ serves as a sanity check. At $k{=}96$ the gap reaches 95 latents: the start view has left every bounded window and must be recalled from memory, making this the discriminative length. HY-WorldPlay posts the strongest baseline scores, but it also moves the least: at $k{=}96$ its median outbound path length is 210\,px, versus 332--834\,px for the other baselines (Table~\ref{tab:memory_base}), and a shorter path is easier to reproduce from memory. Figure~\ref{fig:baseline_qual} shows this regime qualitatively on a $k{=}96$ strafe-and-return rollout: the final chunks must reproduce the input view after a long round trip, and ReWorld returns to the start with the landmark layout intact, whereas the baselines either drift off the commanded path or regenerate the revisited region with altered geometry. The effect of the KV budget and cache policy on our model, including its per-length scaling, is isolated on the long-rollout test in Sec.~\ref{sec:exp-abl} (Table~\ref{tab:ablation}).
% Caveat for the camera-ready pass: confirm the NIAH runs use HY-WorldPlay's flagship autoregressive variant (the 29-frame lightweight variant cannot natively produce a k=96, ~16 s rollout).

\begin{table}[t]
\centering
\caption{Long-horizon memory against the six interactive world-model baselines on the palindromic revisit bench: absolute revisit-similarity scores (SSIM, LPIPS, DINO, ORB) at rollout lengths $k{=}48$ and $k{=}96$ latents, alongside the executed path length (accumulated median optical flow of the outbound half, px). Every method receives the same action commands, but how far it actually travels differs, and a shorter path generally yields higher revisit scores. $k{=}48$ is a sanity length that several baselines can solve from context alone; $k{=}96$ forces recall from beyond every bounded window. Best per column in \textbf{bold}, second best \underline{underlined}; path length is descriptive, not ranked.}
\label{tab:memory_base}
\setlength{\tabcolsep}{4pt}%
\resizebox{\linewidth}{!}{%
\begin{tabular}{l ccccc ccccc}
\toprule
& \multicolumn{5}{c}{$k{=}48$ ($\approx$8\,s)} & \multicolumn{5}{c}{$k{=}96$ ($\approx$16\,s)} \\
\cmidrule(lr){2-6} \cmidrule(lr){7-11}
\rowcolor{stylegreen}
\textcolor{white}{\textbf{Method}} & \textcolor{white}{\textbf{Path Length}} & \textcolor{white}{\textbf{SSIM$\uparrow$}} & \textcolor{white}{\textbf{LPIPS$\downarrow$}} & \textcolor{white}{\textbf{DINO$\uparrow$}} & \textcolor{white}{\textbf{ORB$\uparrow$}} & \textcolor{white}{\textbf{Path Length}} & \textcolor{white}{\textbf{SSIM$\uparrow$}} & \textcolor{white}{\textbf{LPIPS$\downarrow$}} & \textcolor{white}{\textbf{DINO$\uparrow$}} & \textcolor{white}{\textbf{ORB$\uparrow$}} \\
\midrule
SANA-WM~\citep{sanawm2026} & 296 & 0.318 & 0.433 & 0.810 & 0.174 & 499 & 0.313 & 0.528 & 0.700 & 0.161 \\
DreamX~\citep{dreamx2026} & 279 & 0.214 & 0.553 & 0.740 & 0.180 & 454 & 0.194 & 0.627 & 0.603 & 0.159 \\
HY-WorldPlay~\citep{worldplay2025} & 113$^\dagger$ & \textbf{0.448} & \textbf{0.163} & \textbf{0.969} & \textbf{0.464} & 210$^\dagger$ & \textbf{0.427} & \textbf{0.247} & \textbf{0.942} & \textbf{0.460} \\
Matrix-Game 3.0~\citep{matrixgame2026} & 312 & 0.318 & 0.369 & 0.891 & 0.292 & 724 & 0.275 & 0.478 & 0.850 & 0.247 \\
LingBot-World~\citep{lingbotworld2026} & 352 & 0.274 & 0.557 & 0.663 & 0.177 & 834 & 0.251 & 0.635 & 0.509 & 0.162 \\
Yume-1.5~\citep{yume15} & 165 & 0.268 & 0.533 & 0.747 & 0.179 & 332 & 0.269 & 0.586 & 0.661 & 0.152 \\
\midrule
\textbf{ReWorld (ours)} & 286 & \underline{0.349} & \underline{0.282} & \underline{0.913} & \underline{0.325} & 615 & \underline{0.384} & \underline{0.332} & \underline{0.932} & \underline{0.379} \\
\bottomrule
\multicolumn{11}{l}{\footnotesize $^\dagger$\,HY-WorldPlay executes by far the shortest paths: 113/210\,px vs.\ 279--352/454--834\,px for all other methods.}
\end{tabular}}
\end{table}

\subsection{Generation Quality}
\label{sec:exp-vbench}

We assess perceptual quality on the seven video-intrinsic VBench~\citep{vbench} dimensions that require no text prompt: Imaging Quality, Aesthetic Quality, Subject Consistency, Background Consistency, Temporal Flickering, Dynamic Degree, and Motion Smoothness, computed on the same clip set as the controllability benchmark. Native outputs differ across methods in length, resolution, and frame rate, so all clips are normalized to a common specification before scoring---32 uniformly sampled frames spanning the full clip, resized to $1280\times704$, re-encoded at 16\,fps; the comparison is therefore internally controlled, but not directly comparable to numbers reported on native-length outputs. For HY-WorldPlay, this benchmark evaluates its flagship autoregressive model.

\begin{table}[t]
\centering
\caption{VBench quality on the seven video-intrinsic dimensions (higher is better), computed on the shared benchmark clips normalized to 32 frames / $1280\times704$ / 16\,fps. Best per column in bold.}
\label{tab:vbench}
\resizebox{\linewidth}{!}{%
\begin{tabular}{l c ccccccc}
\toprule
\rowcolor{stylegreen}
\textcolor{white}{\textbf{Method}} & \textcolor{white}{\textbf{Mean $\uparrow$}} & \textcolor{white}{\textbf{Imaging}} & \textcolor{white}{\textbf{Aesthetic}} & \textcolor{white}{\textbf{Subject Cons.}} & \textcolor{white}{\textbf{Background Cons.}} & \textcolor{white}{\textbf{Temporal Flick.}} & \textcolor{white}{\textbf{Dynamic Degree}} & \textcolor{white}{\textbf{Motion Smooth.}} \\
\midrule
SANA-WM~\citep{sanawm2026} & 0.835 & 0.668 & 0.537 & 0.908 & 0.919 & 0.942 & 0.900 & 0.969 \\
DreamX~\citep{dreamx2026} & 0.828 & 0.670 & 0.542 & 0.880 & 0.904 & 0.909 & 0.946 & 0.946 \\
HY-WorldPlay~\citep{worldplay2025} & 0.842 & 0.670 & \textbf{0.601} & \textbf{0.976} & \textbf{0.957} & \textbf{0.973} & 0.733 & \textbf{0.987} \\
Matrix-Game 3.0~\citep{matrixgame2026} & 0.836 & \textbf{0.713} & 0.461 & 0.894 & 0.919 & 0.921 & 0.983 & 0.962 \\
LingBot-World~\citep{lingbotworld2026} & 0.841 & 0.692 & 0.577 & 0.905 & 0.915 & 0.918 & 0.933 & 0.947 \\
Yume-1.5~\citep{yume15} & 0.844 & 0.669 & 0.565 & 0.878 & 0.911 & 0.934 & \textbf{0.988} & 0.964 \\
\midrule
\textbf{ReWorld (ours)} & \textbf{0.850} & 0.665 & 0.579 & 0.929 & 0.929 & 0.952 & 0.912 & 0.979 \\
\bottomrule
\end{tabular}}
\end{table}

Table~\ref{tab:vbench} summarizes the seven quality dimensions. ReWorld attains the best mean: no single dimension is dominant, but it is in the first tier on the temporal axes (Motion Smoothness, Temporal Flickering) while sustaining a high Dynamic Degree, whereas the per-dimension leader HY-WorldPlay pays for its consistency scores with markedly lower motion. We note that Dynamic Degree should be read jointly with the revisit metrics of Sec.~\ref{sec:exp-niah}, as consistency and revisit metrics alike favor low-motion rollouts.
% Caveats for the camera-ready pass: (1) HY-AR's low Dynamic Degree (0.733) needs re-verification with matched pose motion magnitude before final claims -- it may partly reflect a setup difference rather than the model. (2) Our weaker dimensions are Imaging Quality (4-step distillation loses fine detail) and Dynamic Degree (output leans stable). (3) With HY's lightweight Wan variant instead of the AR flagship, HY's mean would be 0.874 (first); the flagship is the variant reported in their paper.

\subsection{Ablations}
\label{sec:exp-abl}

\begin{figure}[t]
\centering
\includegraphics[width=\linewidth]{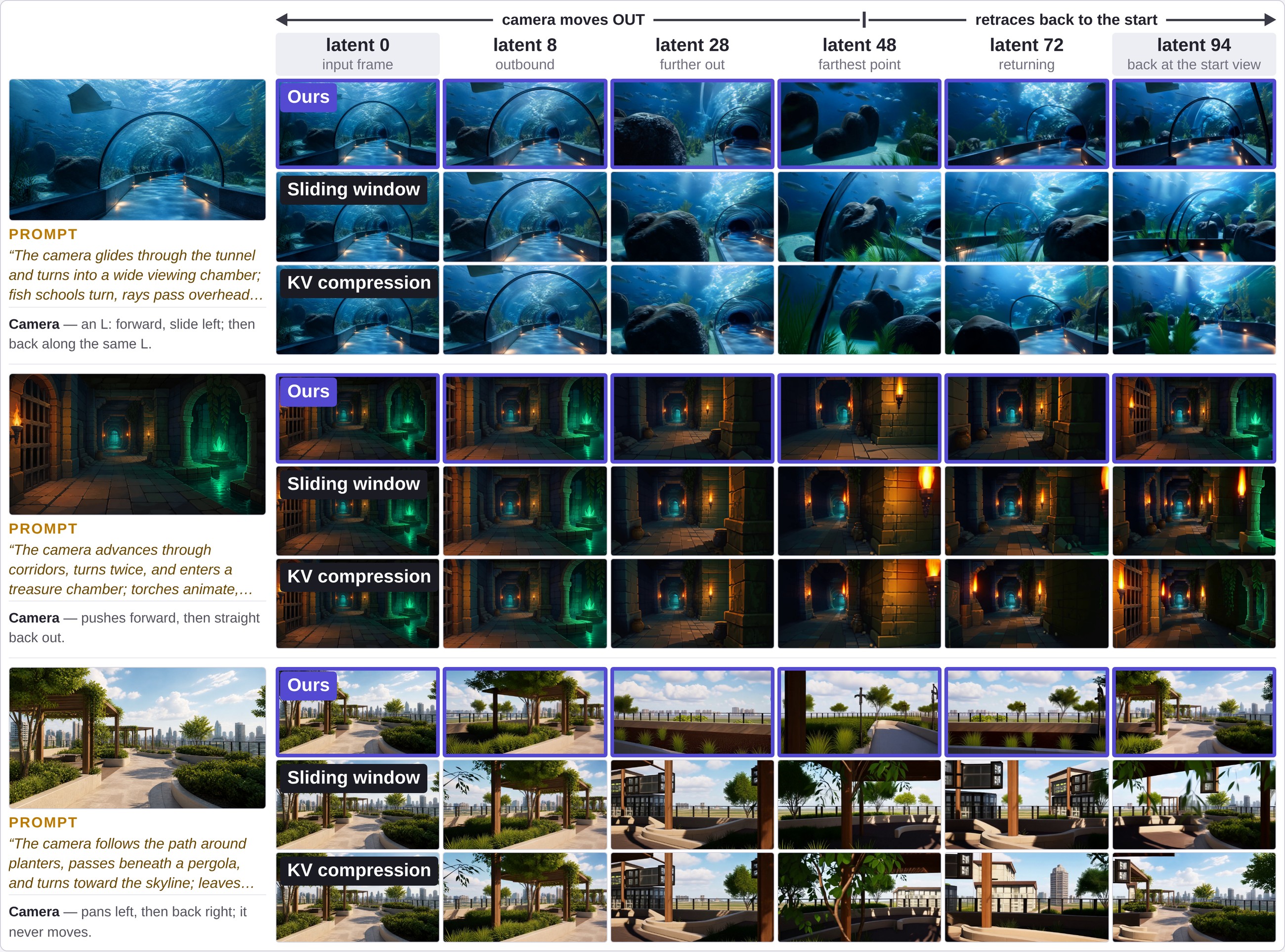}
\caption{\textbf{Inference cache policies, qualitatively.} On out-and-back trajectories, sliding window and KV compression drift or reinvent the revisited scene; the landmark cache (ours) returns to the start view.}
\label{fig:cache_qual}
\end{figure}

\paragraph{Training recipe and cache policy.}
Table~\ref{tab:ablation} ablates two axes on the long-rollout memory test of Sec.~\ref{sec:exp-niah}, reporting revisit SSIM at rollout lengths $k \in \{96,192,288,384\}$ latents. 
The \emph{upper block} varies the training recipe---chunk drop only (CD), random head routing only (RH), and both combined (ours). All variants are trained from scratch under identical settings (720p, 20k steps) and evaluated with the same consolidation arm, so only the recipe differs. 
The \emph{lower block} fixes the final recipe and varies only the inference cache policy under the same budget $B{=}12$. Every bounded arm keeps the same sink chunk and differs only in how the rest of the budget is filled: \emph{window} uses a sliding window of recent chunks; \emph{naive-merge} pools aged chunks instead of evicting them; \emph{consolidation w/o bank} adds six static landmarks (no retrieval); \emph{consolidation w/ bank} retrieves the six landmarks nearest the current pose from the bounded bank (Sec.~\ref{sec:boundedkv}); \emph{full-KV} keeps everything and is the undeployable upper bound (OOM beyond $k{=}192$). The far-gap lengths $k \in \{288, 384\}$, where a sliding window has necessarily evicted the needle, are the discriminative setting; Fig.~\ref{fig:cache_qual} shows the contrast qualitatively.

\paragraph{Control--memory decoupling.}
Table~\ref{tab:fusion} tests whether the routed head structure decouples action-following from long-range recall, comparing three fusion variants trained under the same protocol. \emph{MRoPE only} drops the action embedding and sets all $H{=}24$ heads global, so control must be inferred solely from pose-indexed positional encodings; \emph{Action + MRoPE} keeps the additive action embedding with all heads global, i.e., fusion without routing; \emph{Routing} is our full configuration with the mixed global/local head structure and per-step head routing. We report the control metrics of Sec.~\ref{sec:exp-camctrl} on the long-rollout memory test. Adding action injection improves every control metric over MRoPE only (RotErr $17.66^\circ\!\rightarrow\!13.21^\circ$, TransErr $0.204\!\rightarrow\!0.107$, CamMC $0.394\!\rightarrow\!0.371$) but drops revisit SSIM from 0.3898 to 0.3376---the interference that motivates the window split. Routing matches the unrouted variant on control ($12.94^\circ$, 0.093, 0.346) while restoring revisit SSIM to 0.3752: the structural constraint costs no control authority and preserves long-gap recall.

\FloatBarrier

\section{Related Works}
\label{sec:related}

\paragraph{Streaming video generation.}
\begin{wraptable}{r}{0.5\textwidth}
\centering
\footnotesize
\setlength{\tabcolsep}{3pt}
\renewcommand{\arraystretch}{0.95}
\caption{Revisit SSIM$\uparrow$ on the long-rollout memory test (Sec.~\ref{sec:exp-niah}) at four rollout lengths $k$. \emph{Upper block:} the training recipe is varied with the inference arm fixed to our consolidation. \emph{Lower block:} the inference cache policy is varied on the final recipe; full-KV is the unbounded upper bound (OOM beyond $k{=}192$).}
\label{tab:ablation}
\begin{tabular}{l cccc}
\toprule
& \multicolumn{4}{c}{Revisit SSIM$\uparrow$} \\
\cmidrule(lr){2-5}
\rowcolor{stylegreen}
\textcolor{white}{\textbf{Method}} & \textcolor{white}{\textbf{$k{=}96$}} & \textcolor{white}{\textbf{$k{=}192$}} & \textcolor{white}{\textbf{$k{=}288$}} & \textcolor{white}{\textbf{$k{=}384$}} \\
\midrule
\rowcolor{gray!15}
\multicolumn{5}{l}{\emph{Training recipe (consolidation arm)}} \\
base & 0.4019 & 0.3747 & 0.3698 & 0.3387 \\
+\,CD & 0.4043 & \textbf{0.3941}  & 0.3846 & 0.3463 \\
+\,RH & 0.4319 & 0.3787 & 0.3836 & 0.3565 \\
\textbf{+\,CD\,+\,RH (ours)} & \textbf{0.4358} & 0.3929 & \textbf{0.3943} & \textbf{0.3752} \\
\midrule
\rowcolor{gray!15}
\multicolumn{5}{l}{\emph{Inference cache policy}} \\
\textcolor{gray}{full-KV} & \textcolor{gray}{0.4463} & \textcolor{gray}{0.4231} & \textcolor{gray}{(OOM)} & \textcolor{gray}{(OOM)} \\
window & 0.4129 & 0.3687 & 0.3505 & 0.3476 \\
naive-merge & 0.3741 & 0.3371 & 0.3467 & 0.3541 \\
consolidation w/o bank & \textbf{0.4376} & 0.3758 & 0.3743 &  0.3629\\
\textbf{consolidation w/ bank} & 0.4358 & \textbf{0.3929} & \textbf{0.3943} & \textbf{0.3752} \\
\bottomrule
\end{tabular}
\vspace{10pt}
\centering
\small
\setlength{\tabcolsep}{4pt}
\caption{Control--memory decoupling study: control metrics on the long-rollout memory test (Sec.~\ref{sec:exp-niah}) for the three fusion variants.}
\label{tab:fusion}
\resizebox{\linewidth}{!}{%
\begin{tabular}{l ccc c}
\toprule
\rowcolor{stylegreen}
\textcolor{white}{\textbf{Method}} & \textcolor{white}{\textbf{RotErr$^\circ\downarrow$}} & \textcolor{white}{\textbf{TransErr$\downarrow$}} & \textcolor{white}{\textbf{CamMC$\downarrow$}} & \textcolor{white}{\textbf{Revisit SSIM$\uparrow$}} \\
\midrule
MRoPE only & 17.66 & 0.204 & 0.394 & \textbf{0.3898} \\
Action + MRoPE & 13.21 & 0.107 & 0.371 & 0.3376 \\
\textbf{Routing (ours)} & \textbf{12.94} & \textbf{0.093} & \textbf{0.346} & 0.3752 \\
\bottomrule
\end{tabular}}
\end{wraptable}
Large video diffusion models denoise a clip as a single bidirectional block \citep{wan2025wan, kong2024hunyuanvideo, yang2025cogvideox}, an interface at odds with interaction: no frame can appear before the whole clip is finished. Autoregressive reformulations restore a temporal arrow. Per-frame noise schedules \citep{chen2024diffusionforcing}, chunked causal attention with KV caching \citep{teng2025magi}, context packing \citep{zhang2025packing}, and streaming pipelines \citep{kodaira2026streamdit} realize it, while few-step distillation \citep{yin2024onestepdiffusiondistributionmatching, yin2024improveddistributionmatchingdistillation, wang2024phased, lv2025dualexpertconsistencymodelefficient}, training-free attention reuse \citep{zhao2025real}, self-rollout training that closes the exposure gap of AR \citep{huang2025selfforcing, liu2025rollingforcing}, and AR-train-then-distill systems scaled to long video \citep{longlive2026} make the streams real-time. 
\model follows this recipe---chunked causal generation over a bidirectional backbone, DMD distillation with self-forcing rollouts---but confines distillation to a LoRA \citep{hu2022lora}, so one backbone retains a high-fidelity multi-step mode beside the real-time one.

\paragraph{Interactive world models.}
From recurrent latent simulators \citep{ha2018worldmodels} through latent-action pretraining \citep{bruce2024genie} and its frontier successors \citep{genie3}, interactive generation now spans action-conditioned diffusion trained on game and UE footage \citep{matrixgame2026, lingbotworld2026, sanawm2026, huang2025vid2world}, explorable scene generation \citep{yume15, team2025hunyuanworld, cheng2026360explorer, realwonder2026}, and low-latency streamed deployments \citep{chen2025midas, dreamx2026}. The two prevailing control signals differ in kind: relative pose folded into attention localizes past content \citep{dreamx2026, remind2026}, whereas an injected action supplies the motion command directly \citep{matrixgame2026}, and some recent systems carry both \citep{worldplay2025}. \model likewise keeps both channels, but treats their interference as a training problem, splitting the window each attention head trains under so that control is learned within short windows and memory under long ones.

\paragraph{Memory in world models.}
Long rollouts stay consistent only if views that have left the context remain recallable \citep{gu2025far}. Existing designs differ chiefly in what they store and what that costs: archives of past frames that grow with the rollout, from which pose-relevant views are pulled back as conditioning \citep{yu2025context, worldmem}; an explicit 3D reconstruction that keeps memory outside the generator's own representation \citep{spatialmem2025}; or pose-dependent attention over the retained history \citep{remind2026}. Sinks and sliding windows \citep{streamingllm} bound the cache instead, at the price of forgetting everything beyond the window. \model takes the bounded route without the forgetting: chunks aging out of the recent window consolidate into a fixed-capacity landmark bank inside the model's own KV space, redundancy-based eviction keeps the bank diverse, pose-proximity retrieval fills the cache, and chunk-drop training teaches the model to read the spliced result.

\section{Conclusion}
\label{sec:conclusion}

We presented \model, an interactive streaming world model built in two steps: split the training of control and memory by window---mixed per-head attention windows with random head routing---then consolidate memory at inference, where a bounded cache backed by a pose-indexed landmark bank holds the entire past under a fixed KV budget and chunk-drop training makes its sparse caches in-distribution. Trained on metrically aligned multi-source data and distilled to a few-step LoRA student, \model streams high-resolution video in real time and leads recent interactive world models on control fidelity and visual quality, while revisit fidelity persists at rollout lengths where a sliding window has long evicted the evidence. Memory is still keyed on camera pose alone; extending consolidation to dynamic scenes and richer, non-navigational interaction is the natural next step.

\clearpage
\newpage
\bibliographystyle{style/plainnat}
\bibliography{paper}

@inproceedings{peebles2023scalable,
  title     = {Scalable diffusion models with transformers},
  author    = {Peebles, William and Xie, Saining},
  booktitle = {ICCV},
  pages     = {4195--4205},
  year      = {2023}
}

@article{esser2024scaling,
  title   = {Scaling Rectified Flow Transformers for High-Resolution Image Synthesis},
  author  = {Esser, Patrick and Kulal, Sumith and Blattmann, Andreas and Entezari, Rahim and Müller, Jonas and Saini, Harry and Levi, Yam and Lorenz, Dominik and Sauer, Axel and Boesel, Frederic and Podell, Dustin and Dockhorn, Tim and English, Zion and Lacey, Kyle and Goodwin, Alex and Marek, Yannik and Rombach, Robin},
  journal = {arXiv preprint arXiv:2403.03206},
  year    = {2024}
}

@inproceedings{yang2025cogvideox,
  title     = {CogVideoX: Text-to-Video Diffusion Models with An Expert Transformer},
  author    = {Yang, Zhuoyi and Teng, Jiayan and Zheng, Wendi and Ding, Ming and Huang, Shiyu and Xu, Jiazheng and Yang, Yuanming and Hong, Wenyi and Zhang, Xiaohan and Feng, Guanyu and others},
  booktitle = {ICLR},
  year      = {2025}
}

@article{wan2025wan,
  title   = {Wan: Open and advanced large-scale video generative models},
  author  = {Wan, Team and Wang, Ang and Ai, Baole and Wen, Bin and Mao, Chaojie and Xie, Chen-Wei and Chen, Di and Yu, Feiwu and Zhao, Haiming and Yang, Jianxiao and others},
  journal = {arXiv preprint arXiv:2503.20314},
  year    = {2025}
}

@article{kong2024hunyuanvideo,
  title   = {Hunyuanvideo: A systematic framework for large video generative models},
  author  = {Kong, Weijie and Tian, Qi and Zhang, Zijian and Min, Rox and Dai, Zuozhuo and Zhou, Jin and Xiong, Jiangfeng and Li, Xin and Wu, Bo and Zhang, Jianwei and others},
  journal = {arXiv preprint arXiv:2412.03603},
  year    = {2024}
}

@article{zhang2025packing,
  title   = {Packing input frame context in next-frame prediction models for video generation},
  author  = {Zhang, Lvmin and Agrawala, Maneesh},
  journal = {arXiv preprint arXiv:2504.12626},
  year    = {2025}
}

@article{teng2025magi,
  title   = {MAGI-1: Autoregressive Video Generation at Scale},
  author  = {Teng, Hansi and Jia, Hongyu and Sun, Lei and Li, Lingzhi and Li, Maolin and Tang, Mingqiu and Han, Shuai and Zhang, Tianning and Zhang, WQ and Luo, Weifeng and others},
  journal = {arXiv preprint arXiv:2505.13211},
  year    = {2025}
}

@article{chen2025midas,
  title={Midas: Multimodal interactive digital-human synthesis via real-time autoregressive video generation},
  author={Chen, Ming and Cui, Liyuan and Zhang, Wenyuan and Zhang, Haoxian and Zhou, Yan and Li, Xiaohan and Tang, Songlin and Liu, Jiwen and Liao, Borui and Chen, Hejia and others},
  journal={arXiv preprint arXiv:2508.19320},
  year={2025}
}

@inproceedings{zhao2025real,
  title     = {Real-Time Video Generation with Pyramid Attention Broadcast},
  author    = {Zhao, Xuanlei and Jin, Xiaolong and Wang, Kai and You, Yang},
  booktitle = {ICLR},
  year      = {2025}
}

@article{gu2025far,
  title   = {Long-context autoregressive video modeling with next-frame prediction},
  author  = {Gu, Yuchao and Mao, Weijia and Shou, Mike Zheng},
  journal = {arXiv preprint arXiv:2503.19325},
  year    = {2025}
}

@article{yu2025context,
  title   = {Context as memory: Scene-consistent interactive long video generation with memory retrieval},
  author  = {Yu, Jiwen and Bai, Jianhong and Qin, Yiran and Liu, Quande and Wang, Xintao and Wan, Pengfei and Zhang, Di and Liu, Xihui},
  journal = {ICCV},
  year    = {2025}
}

@article{team2025hunyuanworld,
  title   = {HunyuanWorld 1.0: Generating Immersive, Explorable, and Interactive 3D Worlds from Words or Pixels},
  author  = {Team, HunyuanWorld and Wang, Zhenwei and Liu, Yuhao and Wu, Junta and Gu, Zixiao and Wang, Haoyuan and Zuo, Xuhui and Huang, Tianyu and Li, Wenhuan and Zhang, Sheng and others},
  journal = {arXiv preprint arXiv:2507.21809},
  year    = {2025}
}

@article{huang2025vid2world,
  title   = {Vid2World: Crafting Video Diffusion Models to Interactive World Models},
  author  = {Huang, Siqiao and Wu, Jialong and Zhou, Qixing and Miao, Shangchen and Long, Mingsheng},
  journal = {arXiv preprint arXiv: 2505.14357},
  year    = {2025}
}

@article{bai2025qwen2.5vl,
  title   = {Qwen2.5-vl technical report},
  author  = {Bai, Shuai and Chen, Keqin and Liu, Xuejing and Wang, Jialin and Ge, Wenbin and Song, Sibo and Dang, Kai and Wang, Peng and Wang, Shijie and Tang, Jun and others},
  journal = {arXiv preprint arXiv:2502.13923},
  year    = {2025}
}

@article{wang2024phased,
  title   = {Phased Consistency Model},
  author  = {Wang, Fu-Yun and Huang, Zhaoyang and Bergman, Alexander William and Shen, Dazhong and Gao, Peng and Lingelbach, Michael and Sun, Keqiang and Bian, Weikang and Song, Guanglu and Liu, Yu and others},
  journal = {arXiv preprint arXiv:2405.18407},
  year    = {2024}
}

@article{yin2024onestepdiffusiondistributionmatching,
  title   = {One-step Diffusion with Distribution Matching Distillation},
  author  = {Tianwei Yin and Michaël Gharbi and Richard Zhang and Eli Shechtman and Fredo Durand and William T. Freeman and Taesung Park},
  year    = {2024},
  journal = {arXiv preprint arXiv:2311.18828}
}

@article{yin2024improveddistributionmatchingdistillation,
  title   = {Improved Distribution Matching Distillation for Fast Image Synthesis},
  author  = {Tianwei Yin and Michaël Gharbi and Taesung Park and Richard Zhang and Eli Shechtman and Fredo Durand and William T. Freeman},
  journal = {arXiv preprint arXiv:2405.14867},
  year    = {2024}
}

@article{lv2025dualexpertconsistencymodelefficient,
  title   = {Dual-Expert Consistency Model for Efficient and High-Quality Video Generation},
  author  = {Zhengyao Lv and Chenyang Si and Tianlin Pan and Zhaoxi Chen and Kwan-Yee K. Wong and Yu Qiao and Ziwei Liu},
  year    = {2025},
  journal = {https://arxiv.org/abs/2506.03123}
}

@inproceedings{lipman2023flow,
  title     = {Flow Matching for Generative Modeling},
  author    = {Lipman, Yaron and Chen, Ricky T. Q. and Ben-Hamu, Heli and Nickel, Maximilian and Le, Matt},
  booktitle = {International Conference on Learning Representations},
  year      = {2023}
}

@inproceedings{chen2024diffusionforcing,
  title     = {Diffusion Forcing: Next-Token Prediction Meets Full-Sequence Diffusion},
  author    = {Chen, Boyuan and Marti Mons{\'o}, Diego and Du, Yilun and Simchowitz, Max and Tedrake, Russ and Sitzmann, Vincent},
  booktitle = {Advances in Neural Information Processing Systems},
  year      = {2024}
}

@article{huang2025selfforcing,
  title   = {Self Forcing: Bridging the Train-Test Gap in Autoregressive Video Diffusion},
  author  = {Huang, Xun and Li, Zhengqi and He, Guande and Zhou, Mingyuan and Shechtman, Eli},
  journal = {arXiv preprint arXiv:2506.08009},
  year    = {2025}
}

@article{liu2025rollingforcing,
  title   = {Rolling Forcing: Autoregressive Long Video Diffusion in Real Time},
  author  = {Liu, Kunhao and Hu, Wenbo and Xu, Jiale and Shan, Ying and Lu, Shijian},
  journal = {arXiv preprint arXiv:2509.25161},
  year    = {2025}
}

@article{worldplay2025,
  title   = {WorldPlay: Towards Long-Term Geometric Consistency for Real-Time Interactive World Modeling},
  author  = {Sun, Wenqiang and Zhang, Haiyu and Wang, Haoyuan and Wu, Junta and Wang, Zehan and Wang, Zhenwei and Wang, Yunhong and Zhang, Jun and Wang, Tengfei and Guo, Chunchao},
  journal = {arXiv preprint arXiv:2512.14614},
  year    = {2025}
}

@article{remind2026,
  title   = {Teaching Video Generators to Remember: Eliciting Dynamic Memory for Out-of-Sight State Evolution},
  author  = {Xu, Tianshuo and Xie, Yichen and Meng, Depu and Peng, Chensheng and Herau, Quentin and Jiang, Bo and Hu, Yihan and Zhan, Wei},
  journal = {arXiv preprint arXiv:2605.25333},
  year    = {2026}
}

@article{matrixgame2026,
  title   = {Matrix-Game 3.0: Real-Time and Streaming Interactive World Model with Long-Horizon Memory},
  author  = {Wang, Zile and Liu, Zexiang and Li, Jiaxing and Huang, Kaichen and Xu, Baixin and Kang, Fei and An, Mengyin and others},
  journal = {arXiv preprint arXiv:2604.08995},
  year    = {2026}
}

@article{realwonder2026,
  title   = {RealWonder: Real-Time Physical Action-Conditioned Video Generation},
  author  = {Liu, Wei and Chen, Ziyu and Li, Zizhang and Wang, Yue and Yu, Hong-Xing and Wu, Jiajun},
  journal = {arXiv preprint arXiv:2603.05449},
  year    = {2026}
}

@article{lingbotworld2026,
  title   = {Advancing Open-source World Models},
  author  = {{Robbyant Team} and Gao, Zelin and Wang, Qiuyu and Zeng, Yanhong and Zhu, Jiapeng and Cheng, Ka Leong and Li, Yixuan and Wang, Hanlin and Xu, Yinghao and others},
  journal = {arXiv preprint arXiv:2601.20540},
  year    = {2026}
}

@inproceedings{kodaira2026streamdit,
  title={Streamdit: Real-time streaming text-to-video generation},
  author={Kodaira, Akio and Hou, Tingbo and Hou, Ji and Georgopoulos, Markos and Juefei-Xu, Felix and Tomizuka, Masayoshi and Zhao, Yue},
  booktitle={Proceedings of the IEEE/CVF Conference on Computer Vision and Pattern Recognition},
  pages={29200--29210},
  year={2026}
}

@inproceedings{cheng2026360explorer,
  title={360Explorer: Exploring 4D Controllable World in Panoramic Videos},
  author={Cheng, Xinhua and Zhou, Haiyang and Yu, Wangbo and Jia, Tanghui and Lin, Bin and Ge, Yunyang and Li, Weiqi and Yuan, Li},
  booktitle={Proceedings of the AAAI Conference on Artificial Intelligence},
  volume={40},
  number={5},
  pages={3300--3308},
  year={2026}
}

@inproceedings{streamingllm,
  title     = {Efficient Streaming Language Models with Attention Sinks},
  author    = {Xiao, Guangxuan and Tian, Yuandong and Chen, Beidi and Han, Song and Lewis, Mike},
  booktitle = {The Twelfth International Conference on Learning Representations (ICLR)},
  year      = {2024}
}

@inproceedings{dl3dv,
  title     = {{DL3DV-10K}: A Large-Scale Scene Dataset for Deep Learning-based {3D} Vision},
  author    = {Ling, Lu and Sheng, Yichen and Tu, Zhi and Zhao, Wentian and Xin, Cheng and Wan, Kun and Yu, Lantao and Guo, Qianyu and Yu, Zixun and Lu, Yawen and others},
  booktitle = {Proceedings of the IEEE/CVF Conference on Computer Vision and Pattern Recognition (CVPR)},
  year      = {2024}
}

@article{realestate10k,
  title   = {Stereo Magnification: Learning View Synthesis using Multiplane Images},
  author  = {Zhou, Tinghui and Tucker, Richard and Flynn, John and Fyffe, Graham and Snavely, Noah},
  journal = {ACM Transactions on Graphics (SIGGRAPH)},
  volume  = {37},
  number  = {4},
  year    = {2018}
}

@article{omniworld,
  title   = {{OmniWorld}: A Multi-Domain and Multi-Modal Dataset for {4D} World Modeling},
  author  = {{InternRobotics Team}},
  journal = {arXiv preprint arXiv:2509.12201},
  year    = {2025}
}

@inproceedings{sekai,
  title     = {Sekai: A Video Dataset towards World Exploration},
  author    = {Li, Zhen and Chen, Chuanhao and Yang, Haiyang and Han, Yang and Zhu, Bo and Zhou, Zhuangzi and others},
  booktitle = {Advances in Neural Information Processing Systems (NeurIPS)},
  year      = {2025},
  note      = {arXiv:2506.15675}
}

@article{vipe,
  title   = {{ViPE}: Video Pose Engine for {3D} Geometric Perception},
  author  = {Huang, Jiahui and Zhou, Qunjie and Rabeti, Hesam and Korovko, Aleksandr and Ling, Huan and Ren, Xuanchi and Shen, Tianchang and Gao, Jun and Slepichev, Dmitry and Lin, Chen-Hsuan and others},
  journal = {arXiv preprint arXiv:2508.10934},
  year    = {2025}
}

@inproceedings{megasam,
  title     = {{MegaSaM}: Accurate, Fast, and Robust Structure and Motion from Casual Dynamic Videos},
  author    = {Li, Zhengqi and Tucker, Richard and Cole, Forrester and Wang, Qianqian and Jin, Linyi and Ye, Vickie and Kanazawa, Angjoo and Holynski, Aleksander and Snavely, Noah},
  booktitle = {Proceedings of the IEEE/CVF Conference on Computer Vision and Pattern Recognition (CVPR)},
  year      = {2025}
}

@article{sanawm2026,
  title   = {{SANA-WM}: Efficient Minute-Scale World Modeling with Hybrid Linear Diffusion Transformer},
  author  = {Zhu, Haoyi and Liu, Haozhe and Zhao, Yuyang and Ye, Tian and Chen, Junsong and Yu, Jincheng and He, Tong and Han, Song and Xie, Enze},
  journal = {arXiv preprint arXiv:2605.15178},
  year    = {2026}
}

@article{dreamx2026,
  title   = {DreamX-World 1.0: A General-Purpose Interactive World Model},
  author  = {{DreamX Team}},
  journal = {arXiv preprint arXiv:2606.16993},
  year    = {2026}
}

@article{yume15,
  title   = {Yume-1.5: A Text-Controlled Interactive World Generation Model},
  author  = {Mao, Xiaofeng and Li, Zhen and Li, Chuanhao and Xu, Xiaojie and Ying, Kaining and He, Tong and Pang, Jiangmiao and Qiao, Yu and Zhang, Kaipeng},
  journal = {arXiv preprint arXiv:2512.22096},
  year    = {2025}
}

@inproceedings{vbench,
  title     = {VBench: Comprehensive Benchmark Suite for Video Generative Models},
  author    = {Huang, Ziqi and He, Yinan and Yu, Jiashuo and Zhang, Fan and Si, Chenyang and Jiang, Yuming and Zhang, Yuanhan and Wu, Tianxing and Jin, Qingyang and Chanpaisit, Nattapol and Wang, Yaohui and Chen, Xinyuan and Wang, Limin and Lin, Dahua and Qiao, Yu and Liu, Ziwei},
  booktitle = {CVPR},
  year      = {2024}
}

@article{spatialmem2025,
  title   = {Video World Models with Long-term Spatial Memory},
  author  = {Wu, Tong and Yang, Shuai and Po, Ryan and Xu, Yinghao and Liu, Ziwei and Lin, Dahua and Wetzstein, Gordon},
  journal = {arXiv preprint arXiv:2506.05284},
  year    = {2025}
}

@article{worldmem,
  title   = {WorldMem: Long-term Consistent World Simulation with Memory},
  author  = {Xiao, Zeqi and Lan, Yushi and Zhou, Yifan and Ouyang, Wenqi and Yang, Shuai and Zeng, Yanhong and Pan, Xingang},
  journal = {arXiv preprint arXiv:2504.12369},
  year    = {2025}
}

@misc{genie3,
  title        = {Genie 3: A New Frontier for World Models},
  author       = {Parker-Holder, Jack and Fruchter, Shlomi},
  year         = {2025},
  howpublished = {Google DeepMind Blog, \url{https://deepmind.google/blog/genie-3-a-new-frontier-for-world-models/}},
  note         = {Announced August 5, 2025}
}

@article{ha2018worldmodels,
  title   = {World Models},
  author  = {Ha, David and Schmidhuber, J{\"u}rgen},
  journal = {arXiv preprint arXiv:1803.10122},
  year    = {2018}
}

@inproceedings{bruce2024genie,
  title     = {Genie: Generative Interactive Environments},
  author    = {Bruce, Jake and Dennis, Michael and Edwards, Ashley and Parker-Holder, Jack and Shi, Yuge and Hughes, Edward and Lai, Matthew and Mavalankar, Aditi and Steigerwald, Richie and Apps, Chris and others},
  booktitle = {International Conference on Machine Learning (ICML)},
  year      = {2024},
  note      = {arXiv:2402.15391}
}

@inproceedings{hu2022lora,
  title     = {{LoRA}: Low-Rank Adaptation of Large Language Models},
  author    = {Hu, Edward J. and Shen, Yelong and Wallis, Phillip and Allen-Zhu, Zeyuan and Li, Yuanzhi and Wang, Shean and Wang, Lu and Chen, Weizhu},
  booktitle = {International Conference on Learning Representations (ICLR)},
  year      = {2022},
  note      = {arXiv:2106.09685}
}

@article{longlive2025,
  title={LongLive: Real-time Interactive Long Video Generation},
  author={Yang, Shuai and Huang, Wei and Chu, Ruihang and Xiao, Yicheng and Zhao, Yuyang and Wang, Xianbang and Li, Muyang and Xie, Enze and Chen, Ying-Cong and Lu, Yao and Han, Song and Chen, Yukang},
  journal={arXiv preprint arXiv:2509.22622},
  year={2025}
}

@article{longlive2026,
  title={LongLive-2.0: An NVFP4 Parallel Infrastructure for Long Video Generation},
  author={Chen, Yukang and Wang, Luozhou and Huang, Wei and Yang, Shuai and Zhang, Bohan and Xiao, Yicheng and Chu, Ruihang and Mao, Weian and Hu, Qixin and Liu, Shaoteng and Zhao, Yuyang and Mao, Huizi and Chen, Ying-Cong and Xie, Enze and Qi, Xiaojuan and Han, Song},
  journal={arXiv preprint arXiv:2605.18739},
  year={2026}
}

\clearpage
\newpage

\appendix
\section*{Appendix}

\section{Data Pipeline Details}
\label{app:data}

This appendix expands Sec.~\ref{sec:data} with per-source construction details, the per-source scale divisors and their audits, and the loader-level sampling and conditioning specifics.

\subsection{Per-Source Construction}
\label{app:data-sources}

\paragraph{UE-rendered fly-throughs (metric anchor).}
Our Unreal Engine rendering pipeline produces camera fly-throughs with exact ground-truth trajectories: each clip directory contains the rendered video, a \texttt{camera.json} with per-frame absolute $4\times 4$ camera-to-world matrices (right-handed, Y-up, centimeters) together with intrinsics, and a \texttt{summary.json} recording the scene and trajectory metadata (plus an overhead trajectory visualization). The full rendering run yields $130{,}986$ raw clips over $337$ purchased and in-house environments (indoor, urban, and landscape scenes), drawn from six trajectory modes planned at generation time --- \emph{local\_explore} ($\sim$29\%), \emph{reveal\_pan} ($\sim$26\%), \emph{aerial} ($\sim$15.5\%), \emph{control} ($\sim$15\%), \emph{wandering} ($\sim$8\%), and \emph{boomerang} ($\sim$6\%) --- each walking at a constant, mode-specific speed ($180$\,cm/s for control, $120$\,cm/s for the exploratory modes). Because raw mode and scene frequencies are skewed, we curate a balanced subset by scene round-robin: for each mode we target $\sim$8{,}000 clips while cycling over environment and level identifiers, which yields $47{,}726$ clips spanning $248$--$324$ environments per mode; darkness filtering (Sec.~\ref{sec:data-sampling}) reduces this to the $42{,}536$-clip expansion set of Table~\ref{tab:data}, whose captions combine the pipeline scene summary with a camera-motion sentence. The $95{,}895$-clip control set of Table~\ref{tab:data} comes from an earlier, control-heavy rendering run ($27{,}551$ renders of $60$\,s, split into $110{,}204$ clips) and is obtained from $109{,}132$ captioned clips by the same darkness filter (removing black-frame and wall-facing clips); its trajectories are generated from the explicit discrete action space of the control mode, with per-frame action labels kept consistent with the realized motion, and its captions are generated by the rendering pipeline itself. Because UE poses are exact and metric, these sources serve as the scale anchor in Sec.~\ref{sec:data-metric}.

\paragraph{UE luminance and motion filtering.}
UE renders fail in a characteristic way: when the camera clips into geometry or faces an unlit surface, frames are near-black, so both UE sets pass through a luminance filter. Each clip is probed with $16$ uniformly sampled grayscale frames and dropped if its mean luminance is below $35$ (on $0$--$255$) or if more than half of the sampled frames are dark; a lightweight motion probe during captioning separately removes frozen clips whose frames barely change. This filter reduces the control set from $109{,}132$ captioned clips to $95{,}895$ ($-12.1\%$) and the round-robin-selected expansion set from $47{,}726$ clips to $42{,}536$ ($-10.9\%$).

\paragraph{DL3DV.}
From DL3DV-10K real-scene walkthroughs \citep{dl3dv} we cut frame-accurate $15$\,s slices whose pose sub-segments are aligned to the video slice with zero drift, yielding $29{,}864$ clips with no missing modality. Poses are estimated with VIPE \citep{vipe} (near-metric). Each clip is captioned with Qwen-VL \citep{bai2025qwen2.5vl} (average $149$ words per caption, no duplicates or empties).

\paragraph{RealEstate10K.}
We use RealEstate10K \citep{realestate10k} indoor/real-estate footage (watermark-inpainted video), retaining clips of at least $8$\,s, which leaves $12{,}065$ of the original $22{,}275$. Rather than the released SLAM trajectories, poses are re-estimated with VIPE so that pose provenance matches DL3DV; captions come from a pre-computed caption store.

\paragraph{Game roaming.}
The game-roaming source is curated from a $494.7$-hour collection of gameplay recordings spanning $168$ games. A roaming whitelist keeps $88$ titles suited to free exploration and walking (pure shooters are excluded), which are sliced into $24{,}801$ clips of $15$\,s at $1920\times1080$, $30$\,fps. Each clip receives VIPE metric camera-to-world poses (OpenCV convention, first frame anchored to the identity) and a Qwen-VL caption; successive cleaning passes remove pose-quality outliers, dark clips, and full-screen menu or loading content, leaving $18{,}387$ clips across $79$ games (nine whitelisted titles yield no surviving clips). Because the VIPE poses of this source are already metric, it enters Sec.~\ref{sec:data-metric} with $\sigma = 1$.

\paragraph{OmniWorld-Game.}
From the simulator/game domain of OmniWorld \citep{omniworld} ($479$ scenes) we normalize the released quaternion extrinsics, invert the world-to-camera matrices to camera-to-world at build time, and merge only frame-contiguous reconstruction splits into runs (splits across index gaps are independent reconstructions --- largely teleports --- and are never merged). Runs are cut into $12$\,s windows with $9$\,s stride ($25\%$ overlap), discarding runs shorter than $8.5$\,s, and each clip is re-anchored so that its first pose is the identity. This yields $5{,}629$ clips (median $12.0$\,s) with per-clip JSON captions.

\paragraph{Sekai real-walking-hq and game-walking.}
From the Sekai corpus \citep{sekai} of first-person walking videos we intersect available $720$p/$60$\,s videos with released pose files, obtaining $14{,}730$ real-walking clips; clips are kept as uncut $60$\,s segments. Poses are per-frame MegaSaM \citep{megasam} camera-to-world estimates (the c2w convention was verified empirically by forward-direction alignment, and pose--video synchronization was verified by cross-correlating optical-flow speed with pose angular speed, with median lag $0$). The companion game-walking split contributes $1{,}618$ clips rendered in UE5 with engine ground-truth c2w trajectories, downscaled to $720$p without re-cutting. Both splits use the official CSV captions (caption plus location, weather, crowd, and time-of-day tags).

\subsection{Alignment Details and Audits}
\label{app:data-alignment}

\paragraph{Per-source divisors.}
The divisor of Eq.~\eqref{eq:transscale} is measured by replaying the exact training-time windowing (a $189$-frame window resampled to $24$\,fps, i.e.\ $0.1640625$\,s per latent step, with all trajectory augmentation disabled), converting poses to per-latent translation increments $\Delta t$, and pooling the increment magnitudes over sampled clips. The resulting divisors are: UE $\sigma=100$ (exact cm$\to$m conversion; anchor), DL3DV $0.0709/0.3667 = 0.1923$, RealEstate10K $0.0825/0.3667 = 0.2273$, OmniWorld-Game $0.641/0.3667 = 1.75$ (pooled over the locomotion subset, since the UE anchor is locomotion-dominated and including the $29\%$ vehicle clips would inflate the divisor), Sekai real-walking $0.0032/0.3667 = 0.0086$, and Sekai game-walking $0.0039/0.3667 = 0.0107$; the game-roaming source is already metric under VIPE and uses $\sigma = 1$. The divisor is applied to the translation column of every pose in the loader. The UE expansion set moves genuinely more slowly than the control set (its raw per-latent-step translation median is $\approx 20$\,cm versus $\approx 37$\,cm); we deliberately do not renormalize it separately, since the speed difference is real rather than a scale error.

\paragraph{Spread and post-hoc audits.}
A single global scalar per source suffices because the per-clip scale spread is bounded: the $p_{90}/p_{10}$ ratio of per-clip median increments is $3.65$ (DL3DV), $2.66$ (RealEstate10K), and $2.86$/$2.67$ (Sekai real/game) --- under one order of magnitude and attributable to genuine capture-speed variation --- so neither per-scene normalization nor metric-depth rescaling is needed. A post-hoc audit of the model-facing increments after rescaling confirms alignment: the per-source median $\lVert \Delta t \rVert$ relative to UE is $1.00$ (UE, by construction), $0.92$ (DL3DV), $1.00$ (RealEstate10K), $1.21$ (OmniWorld-Game, a vehicle-tail effect), $0.96$ (Sekai real), and $0.94$ (Sekai game).

\paragraph{Coordinate-convention normalization.}
Coordinate conventions are normalized jointly with scale. All sources are brought to camera-to-world matrices in the OpenCV camera basis (forward $+z$, down $+y$): the UE camera basis (forward $+x$, up $+y$) is conjugated by a non-trivial $90^{\circ}$ rotation about the vertical axis (a naive $\mathrm{diag}(1,-1,-1)$ sign flip is incorrect and was ruled out by an empirical derivation against engine trajectories); OmniWorld world-to-camera extrinsics are inverted at clip-build time; Sekai extrinsics are already c2w and are ingested without inversion; the legacy camera-referenced (w2c) RealEstate10K mode is not used in the joint configuration. The audit further verifies that forward motion maps to a dominant $+t_z$ in every source (DL3DV is $t_x$-dominated in aggregate, but this reflects its orbital/strafing capture style rather than a convention error) and that yaw maps to the same Euler index with consistent sign. Rotations are already dimensionless (radians) and are never rescaled; only translation passes through Eq.~\eqref{eq:transscale}.

\subsection{Sampling and Conditioning Details}
\label{app:data-sampling}

\paragraph{Window extraction.}
Within a long clip the $189$-frame window --- the native training-window length of $48$ latent frames, i.e.\ $L{=}12$ chunks of $4$ --- starts at a uniformly random frame, and frames are resampled to $24$\,fps by uniform striding (clips shorter than the required span are stretched by linear index interpolation). A $60$\,s Sekai clip therefore contributes a different random $\approx 8$\,s excerpt ($7.88$\,s) on every epoch, which converts long uncut footage into diverse window-level supervision without offline re-slicing.

\paragraph{Palindrome augmentation.}
The palindrome augmentation of Sec.~\ref{sec:data-sampling} forces the camera to retrace its own path within a single training window and additionally provides reverse-motion coverage. Under this augmentation, pose downsampling is SE(3)-aware (rotation slerp), since arithmetic averaging across the palindrome mirror point produces singular matrices. A whole-window reversal probability exists in the loader but is disabled in the joint configuration.

\paragraph{Emitted conditioning.}
From each window the loader emits the RGB frames, the per-chunk actions, and the pose context consumed by MRoPE. The $189$ frame poses are downsampled to the $48$ latent steps, re-anchored so the first camera is the identity, and differenced into per-step 6-DoF increments (three translations plus three Euler angles), which are grouped per latent chunk into the actions $a_k \in \mathbb{R}^6$. The MRoPE context is the relative camera-to-world trajectory $P_k \in \mathrm{SE}(3)$ itself, with translations additionally normalized by the window's maximum radius (a pure rescaling that preserves the identity anchor), so that pose-indexed attention operates on a bounded, source-agnostic coordinate frame.

% \section{Contributions and Acknowledgements}
% \label{app:contributions-and-acknowledgements}

\end{document}